\documentclass{article}

\PassOptionsToPackage{numbers,compress}{natbib}
 \usepackage[preprint]{neurips_2026}

\usepackage{float}
\usepackage{multirow}
\usepackage{booktabs}
\usepackage{makecell}
\usepackage{graphicx}
\usepackage{caption}
\usepackage[utf8]{inputenc} 
\usepackage[T1]{fontenc}    
\usepackage{hyperref}       
\usepackage{url}            
\usepackage{booktabs}       
\usepackage{amsfonts}       
\usepackage{nicefrac}       
\usepackage{microtype}      
\usepackage{xcolor}         
\usepackage{enumitem}
\usepackage{graphicx}
\usepackage{changes}
\usepackage{makecell}
\usepackage{multirow}
\definechangesauthor[name={Yuval}, color=blue]{Y}
\definechangesauthor[name={Mati}, color=purple]{M}
\definechangesauthor[name={Navve}, color=orange]{N}

\title{
From Activation to Specificity:
Automating Counterfactual Testing of Visual Representations in the Human Brain
}

\author{%
  \normalfont
  Yuval Golbari$^{1,*}$ \quad
  Navve Wasserman$^{1,*}$ \quad
  Matias Cosarinsky$^{1}$ \quad
  Roman Beliy$^{1}$ \quad
  Aude Oliva$^{2}$  \\
  Antonio Torralba$^{2}$ \quad
  Michal Irani$^{1}$ \quad
  Tamar Rott Shaham$^{2}$ \\[6pt]
  {\small $^{1}$Weizmann Institute of Science} \quad
  {\small $^{2}$Massachusetts Institute of Technology} \quad
  {\small $^{*}$Equal contribution}
}

\begin{document}

\maketitle

\begin{abstract}

Identifying which brain regions represent a visual concept in the human brain is a central challenge in neuroscience. Existing approaches have localized coarse functional regions (e.g., faces, places) through activation maximization, identifying regions that activate strongly for a target concept relative to other concepts. Yet strong activation alone does not establish that a region represents the concept itself, as responses may instead be driven by correlated visual or semantic cues.
We introduce BrainTRACE (Testing Representations through Counterfactual Evidence), an automated framework that combines generative and brain models to synthesize controlled stimuli and validate neural representations through targeted counterfactual-specificity testing. Given a query specifying a concept of interest, our framework constructs targeted stimulus sets comprising concept images, counterfactual edits that remove the target concept while preserving other image content, and images with candidate
correlated distractors. It then uses an image-to-fMRI encoding model to predict brain responses and searches for representations that respond specifically to the target concept over correlated alternatives. BrainTRACE returns validated candidate representations and proposes follow-up fMRI experiments to further test or extend its discoveries. Our approach successfully recovers known functional localizations and identifies new candidate representations across dozens of concepts, validated on both predicted and measured fMRI data. Critically, we show that without counterfactual evaluation, a large fraction of localizations would be false positives, confirming that activation alone is insufficient evidence of representation.
For the generated stimuli, code, and additional results, see our \href{https://yuvalgol123.github.io/BrainTrace/}{project page}.

\end{abstract}

\section{Introduction}

The human brain organizes visual experience into representations of objects, scenes, actions, and abstract semantic structures. Understanding how such visual concepts are represented in the brain is therefore a central goal of neuroscience. Over the past decades, studies using functional magnetic resonance imaging (fMRI) have revealed important aspects of visual organization, from low-level visual maps~\cite{Sereno1995Science,Engel1997CerebCortex} to general category-selective regions~\cite{Kanwisher1997,Epstein1998,downing2001cortical,cohen2000visual}, while recent computational approaches have begun to discover richer representations~\cite{Bao2025MindSimulator,hwang2025silico,luo2024brain,neurogen,van2025multidimensional}.
At the core of this progress lies a standard methodology measuring category contrasts~\cite{Kanwisher1997,Epstein1998,downing2001cortical,cohen2000visual}, asking whether a brain region or voxel responds more strongly to a target category than to the average response across other categories. Such analyses have led to classical findings of regions selective for general concepts such as faces, bodies, and words. More recently, image-to-fMRI encoders have extended this paradigm by predicting brain activity for images that were never measured, allowing the use of more data and revealing more nuanced representations~\cite{Bao2025MindSimulator,wasserman2025brainexplore}.

Despite this progress, important limitations remain. Many methods do not test whether a strongly responding voxel or region truly represents the target concept, or whether its response is instead driven by correlated visual or semantic cues that co-occur with the concept, such as color, background, pose, or the presence of other objects. Without such counterfactual-specificity testing, many discovered localizations may in fact be false positives. 
Beyond identifying candidate representations, current methods provide limited guidance about which follow-up fMRI experiments are worth running, which concepts are underrepresented in existing datasets, and which new stimuli would be most informative to add.

\begin{figure}[t]
    \centering
    \includegraphics[width=0.98\textwidth]{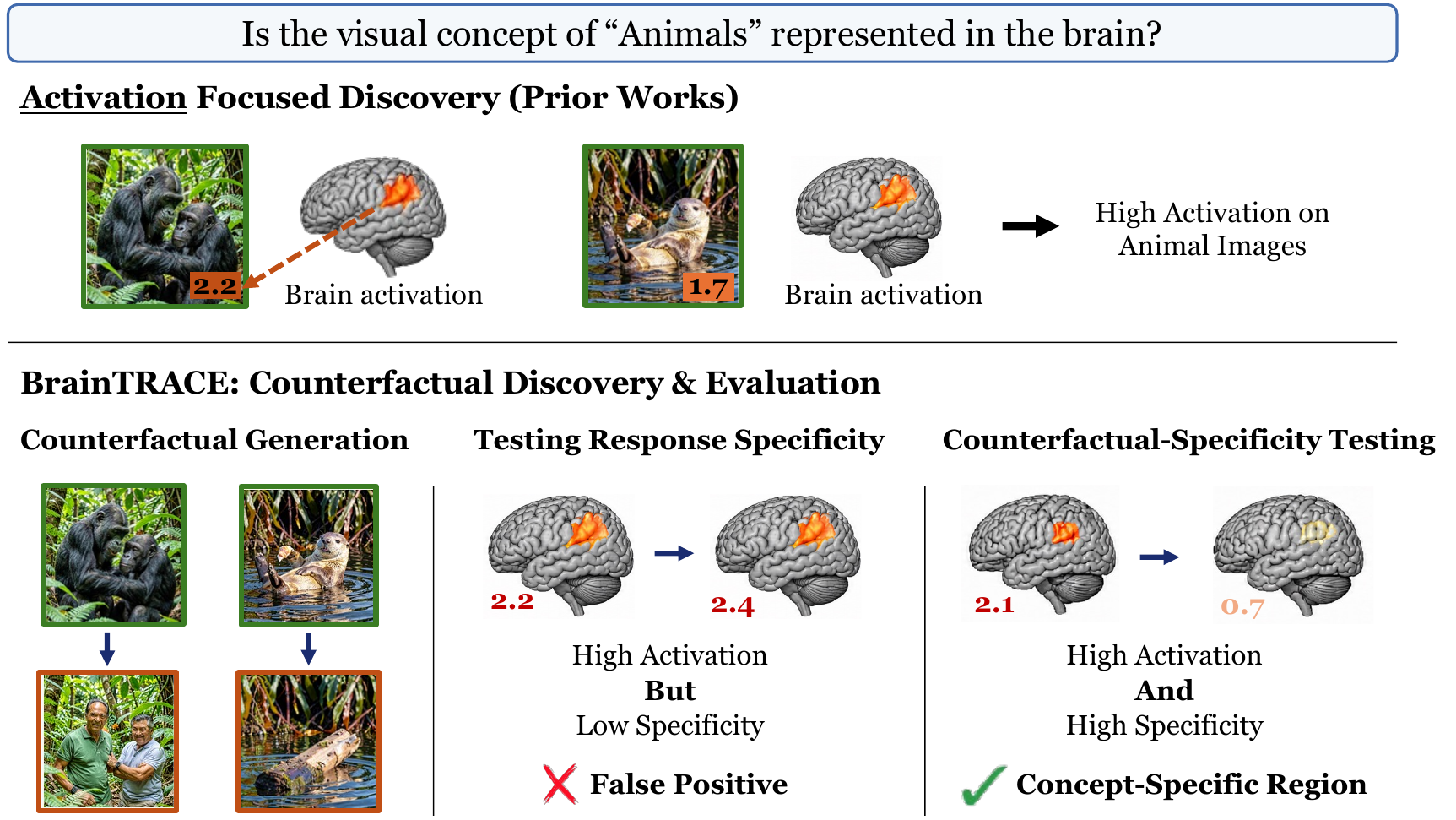}
    \caption{\textbf{Overview of our approach.}
    Activation-maximization based methods (top) identify regions with high responses to a target concept, but cannot distinguish true concept representations from correlated cues. In contrast, BrainTRACE (bottom) performs counterfactual-specificity evaluation using targeted counterfactual stimuli that isolate the concept. Regions with high activation but low response difference between original and counterfactual images are identified as false positives, while regions with both high activation and strong specificity response are identified as candidate concept-specific representations. 
    }
    \vspace{-0.4cm}

    \label{fig:overview}
\end{figure}

To address these limitations, we introduce \emph{BrainTRACE}, an automated framework for discovering and evaluating candidate visual concept representations through counterfactual-specificity testing. Given a target concept, BrainTRACE constructs targeted stimulus sets designed to isolate the concept from correlated visual and semantic factors. It then uses an image-to-fMRI encoding model to predict brain responses and identifies voxels and regions that respond selectively to the concept rather than to co-occurring properties (Fig.~\ref{fig:overview}). When evidence is insufficient, it proposes follow-up experiments to further test its discoveries.

More specifically, the targeted image--fMRI dataset contains three types of images: \emph{\textbf{(i) Positive Images}}, which depict the target concept, obtained through both generation and retrieval from large image pools. \emph{\textbf{(ii) Semantically Negative Images}}, which depict correlated but distinct concepts proposed by a large language model that identifies concepts likely to co-occur with the target but not be the concept itself. For example, for \emph{human hands}, it may propose alternatives such as \emph{human body}, \emph{human legs}, or \emph{robot hands}. \emph{\textbf{(iii) Counterfactual Negative Images}} are created by editing positive images to remove or replace the target concept while preserving the rest of the image content, helping rule out explanations based on background, color, object position, or other co-occurring properties. Generation is particularly useful for providing clear and diverse instances, especially for concepts that are rare or underrepresented in existing datasets. Importantly, when available, BrainTRACE also retrieves positive and semantically negative images from the measured image--fMRI dataset itself, which are used to evaluate the discovered representations.

Together, this enables BrainTRACE to characterize discovered representations in terms of whether a candidate representation responds specifically to the target concept under the tested alternatives, where it is represented, and with what level of confidence. It further assesses how well the concept is supported by the measured data and whether this evidence is sufficient to substantiate the finding. When evidence is insufficient, it proposes stimuli for follow-up fMRI experiments, including positive examples, semantic negatives, and counterfactual edits.

Using BrainTRACE, we study 260 visual concepts. We first show that many representations identified by activation-based methods are false positives under counterfactual-specificity evaluation, with over 70\% failing to exhibit concept-specific responses under counterfactual-specificity evaluation, despite achieving high activation scores. Using specificity scores, BrainTRACE then identifies which concepts are robustly represented and where, evaluated on both predicted and measured fMRI data. We validate the framework against known brain regions, showing it recovers areas selective for faces, bodies, places, and words, before turning to a broader range of representations supported by counterfactual-specificity evidence, including different body parts such as hands and legs; text-related concepts such as handwritten text, traffic signs, and logos; and additional concepts such as animal faces, food, tools, and social interactions.

By moving beyond activation-based analyses toward targeted counterfactual-specificity testing, BrainTRACE not only reveals that many previously identified representations may be driven by correlated factors rather than the concepts themselves, but provides a systematic approach for discovering robust, concept-specific representations. Through large-scale evaluation across hundreds of concepts, we demonstrate that counterfactual-specificity testing leads to more reliable and interpretable mappings between visual concepts and brain activity. Beyond analysis, BrainTRACE closes the loop with experimental design, automatically identifying gaps in existing data and proposing informative stimuli for follow-up studies. In doing so, it offers a unified framework for discovery, validation, and refinement of visual representations in the brain.

\vspace{-0.1cm}
\section{Related Works}
\vspace{-0.25cm}

A foundational goal of visual neuroscience is to understand where and how the brain represents the visual world. Early fMRI studies revealed that visual cortex is organized retinotopically and encodes basic image properties such as orientation, spatial frequency, color, and motion~\cite{Sereno1995Science, Engel1997CerebCortex, DeYoe1996PNAS, Kamitani2005NatNeurosci, Henriksson2008JOV, Conway2007Neuron, Tootell1995Nature}. While these low-level properties are easier to systematically manipulate and disentangle, the space of higher-level visual concepts is vastly larger and less structured. At this higher level, researchers presented subjects with images from a small set of carefully chosen, labeled categories and identified regions responding selectively to each, revealing broad category-selective regions in higher visual cortex for faces~\cite{kanwisher2006fusiform}, scenes~\cite{epstein2019scene}, bodies~\cite{downing2001cortical}, and written words~\cite{cohen2000visual}. While highly influential, this approach was inherently limited to a small, pre-specified set of broad concepts.

More recent work has sought to move beyond these narrow category sets. Advances in brain modeling~\cite{Wasserman2026NoShared, Yamada2015Converter, Scotti2024MindEye2, Beliy2025BrainIT,ratan2021computational}, and in particular encoding models that predict voxel-wise fMRI responses from image features~\cite{Kay08, Naselaris11, Beliy2024Wisdom, adeli2025transformer}, have enabled concept analysis at a much larger scale by allowing predictions for images never measured in the scanner. One line of work uses such models to characterize already-defined regions by retrieving images from large datasets that most strongly activate a given region~\cite{hwang2025silico}, or by generating images that maximize its activation~\cite{neurogen, braindiffusion}. Other recent approaches, such as MindSimulator~\cite{Bao2025MindSimulator}, use concept-oriented image retrieval together with predicted fMRI responses to localize candidate concept-selective voxels. Yet a central limitation remains: a region that responds strongly to images of a concept may still be driven by correlated visual or semantic properties rather than by the concept itself.


Recent advances in brain modeling together with machine learning offer a new path forward. Modern generative models can synthesize and edit images with fine-grained semantic control~\cite{rombach2022high, brooks2023instructpix2pix, wasserman2025paint, labs2025flux1kontextflowmatching}, while large language models can help propose correlated counter-hypotheses and reason about informative interventions~\cite{kiciman2023causal, alkan2025survey, zhou2024hypothesis}. 
Combined with image-to-fMRI encoding models, these capabilities make it possible to construct controlled stimulus sets that disentangle concept-driven activation from correlated factors, and to predict the corresponding fMRI responses. We propose BrainTRACE, which leverages these tools to address the specificity gap in activation-based visual concept representation discovery and to guide future experiments when existing data is insufficient for proper validation.

\begin{figure}[t]
\centering

\includegraphics[width=1\textwidth]{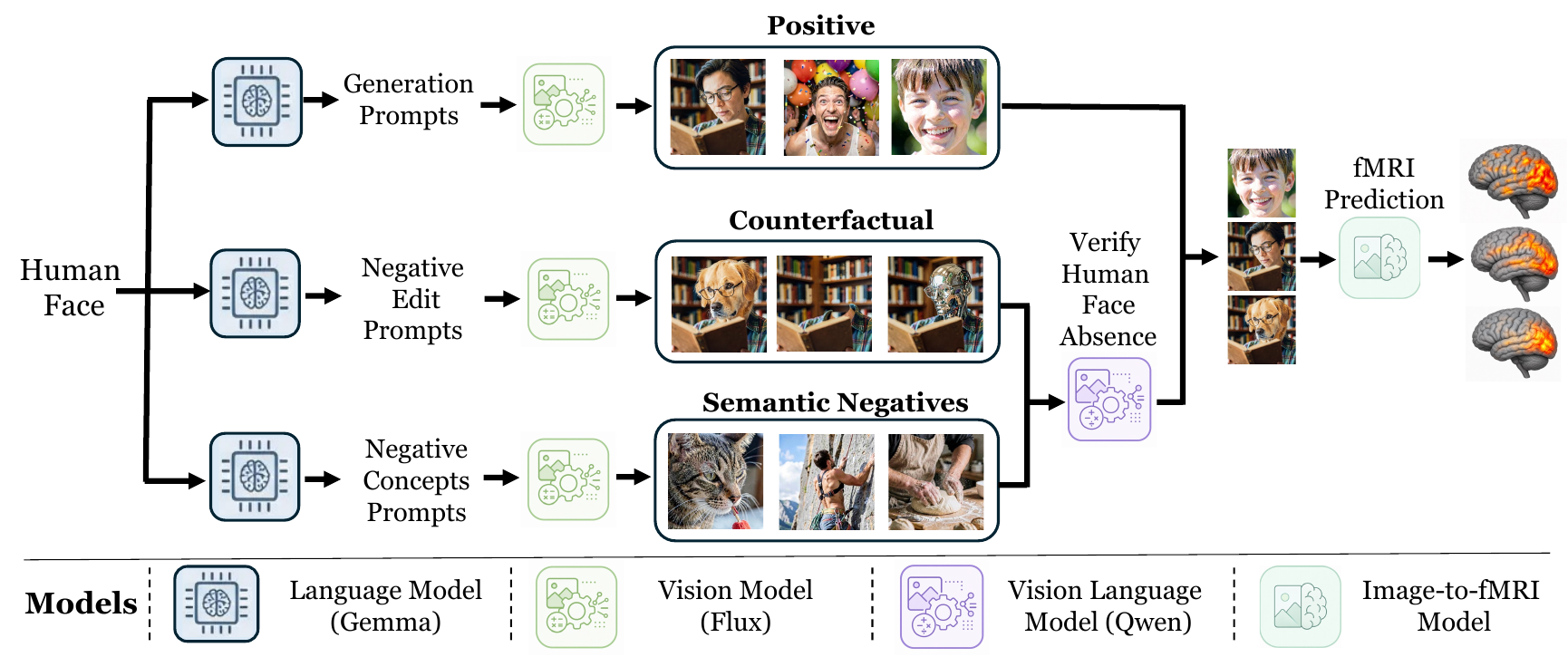}
\caption{\textbf{Concept-Targeted Counterfactual Stimulus Generation.} Given a target concept (e.g., human face), BrainTRACE constructs a targeted stimulus set consisting of three types of stimuli: positive images, counterfactuals, and semantic negatives. A language model generates prompts for each type, which are used by a text-to-image model to synthesize images. Counterfactual and semantic negatives are designed to isolate the target concept from correlated visual and semantic factors. A vision-language model verifies the presence or absence of the target concept in each image. Finally, all images are passed through an image-to-fMRI model to obtain predicted neural responses.}
\label{fig:method}
\vspace{-0.1cm}
\end{figure}

\vspace{-0.08cm}
\section{The BrainTRACE Framework}
\vspace{-0.10cm}

BrainTRACE takes as input a target concept together with a subject's image--fMRI dataset, and outputs a candidate voxel set representing that concept along with a confidence estimate. We automatically constructed a list of 260 concepts using ChatGPT (GPT-5)~\cite{openai2025gpt5systemcard}, with the goal of covering a broad and diverse range of visual concepts.
Each image is associated with an fMRI response vector, where each entry corresponds to the activation of a voxel in the cortex ($\sim$ 40K voxels in our data).
The pipeline proceeds in three stages; \textbf{\textit{(i) targeted stimulus set generation}}. Given a target concept, BrainTRACE constructs a targeted stimulus set designed to isolate the target concept from correlated visual and semantic factors. These include: 
positive images of the concept, semantic negatives depicting correlated but distinct concepts, and counterfactual negatives in which the target concept is removed or replaced while the rest of the image is preserved as much as possible. In addition, it retrieves positive and semantic-negative images from large image pools and, when available, from the measured image--fMRI dataset itself. Images without measured fMRI are passed through an image-to-fMRI encoder to obtain predicted responses. \textbf{\emph{(ii) Concept-selective representation search.}} BrainTRACE then computes activation and specificity scores for each voxel based on the positive, semantic-negative, and counterfactual image sets, and uses these scores to identify a candidate voxel set with strong and concept-specific responses. \textbf{\emph{(iii) Validation and Follow-Up Experiment Design.}} BrainTRACE then validates this candidate set using held-out predicted and measured fMRI data, quantifies how well the concept is represented in the measured dataset, and combines these signals into a final verdict. When measured evidence is insufficient, BrainTRACE analyzes which evidence is missing and proposes informative follow-up stimuli for future fMRI experiments.

\subsection{Concept-Targeted Stimulus Set Generation}
\label{sec:method:Steps:Generation}

In this stage, BrainTRACE constructs a dedicated image--fMRI dataset for the target concept, designed to localize candidate representations while testing their specificity under counterfactual and semantic-negative comparisons. Starting from the target concept, it generates three types of image stimuli, highlighted in Fig.~\ref{fig:method}. \emph{\textbf{(i) Positive Images:}} A large language model (Gemma-3-27B-IT~\cite{gemma_2025}) produces diverse prompts representing the target concept, aiming to capture a wide range of visual variations. These prompts are passed to a text-to-image model (FLUX.2~\cite{flux-2-2025}) to synthesize one image per prompt. For each concept, BrainTRACE generates 200 positive images for training and 100 additional held-out images for validation. \emph{\textbf{(ii) Semantic Negative Images:}} The language model also proposes a set of 10 \emph{counter concepts} --- concepts that are semantically or visually related to the target concept, and may co-occur with it, but do not require the target concept itself to appear. For example, for the target concept \emph{person surfing}, proposed counter concepts may include \emph{a man fishing}, \emph{beach}, or \emph{wave}. For each counter concept, the language model generates 10 prompts while explicitly checking that the target concept is not mentioned in any of them. The text-to-image model synthesizes the corresponding images. An Image--Concept Verification step using a vision-language model (Qwen3-VL-8B~\cite{qwen3technicalreport}) is then applied to verify that the target concept is indeed absent from the images. After filtering, this stage yields approximately 80--100 semantic negative images for training and a similar number for validation. \emph{\textbf{(iii) Counterfactual Negative Images:}} For each selected positive image, the language model proposes 10 edit instructions designed to minimally modify the image so that it no longer expresses the target concept. For example, for the concept \emph{human face}, the edits may replace the human face with an animal face or remove the face entirely. These prompts are passed to an image editing model (FLUX.2), and the resulting edited images are verified. This process is applied to 50 training images and 20 validation images, yielding approximately 400--500 counterfactual negatives. Finally, all generated images are passed through an image-to-fMRI encoder to obtain predicted brain responses.

In addition to generated stimuli, BrainTRACE retrieves positive and semantic-negative images from the measured image--fMRI dataset using a CLIP-based retrieval strategy~\cite{radford2021learning} inspired by~\cite{Bao2025MindSimulator}. Retrieved images are filtered to ensure that the target concept is present in positives, absent from the negatives, and that negatives align with the intended alternative concept. This retrieval stage enables evaluation on measured data for concepts and counter concepts that are sufficiently represented in the dataset. 

\begin{figure}[t]
    \centering
    \includegraphics[width=1.0\textwidth]{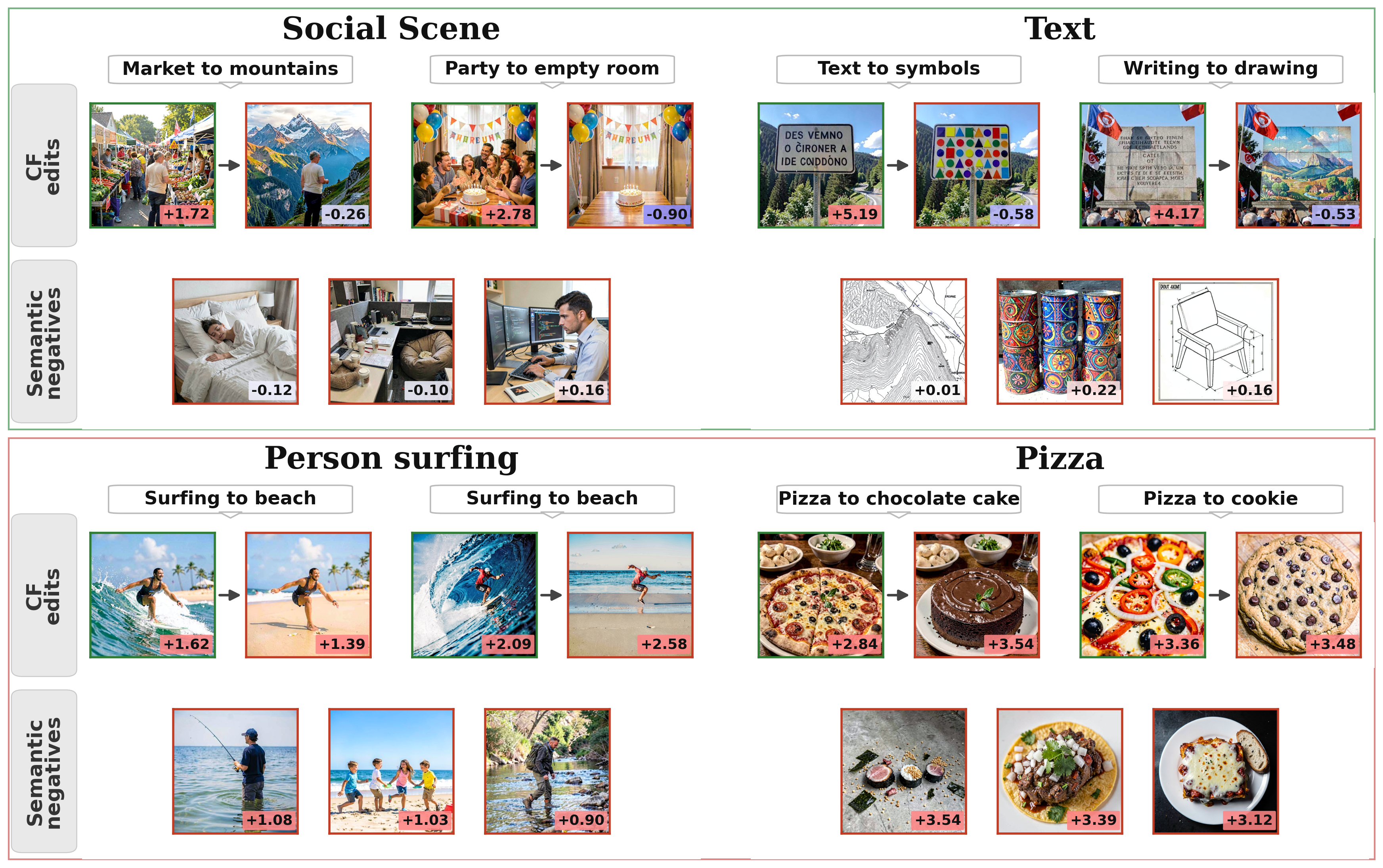}
    \vspace{-0.36cm}
    \caption{\textbf{Counterfactual-specificity evaluation of discovered regions.} Top: regions discovered by BrainTRACE, showing high activation for positives and lower activation for counterfactual edits and semantic negatives. Bottom: regions discovered by an activation-based method, which remain highly activated after edits or for related negatives, indicating false positives driven by correlated cues.}
    \vspace{-0.14cm}
    \label{fig:Good_and_bad_Casulity_Edits}
\end{figure}

\vspace{-0.1cm}
\subsection{Concept-Selective Representation Search}
\label{sec:method:Steps:Search}
\vspace{-0.08cm}

Given the concept-targeted stimulus set, BrainTRACE assigns each voxel three scores with respect to the target concept. The first is an \emph{\textbf{activation score}}, measuring how strongly the voxel responds on average to the positive images. The second calculates the difference between the activation score and the hardest semantic negatives, defined as the 10 negative images that most strongly activate the voxel. The third calculates the difference between each positive image and its hardest edited counterpart, using the edit that produces the highest activation for that voxel. Intuitively, the activation score tests whether a voxel responds strongly to the concept, while the two \emph{\textbf{specificity scores}} test whether this response remains higher than semantically related alternatives and controlled counterfactual edits.

BrainTRACE then uses the specificity scores for each voxel as the average of the semantic-negative and counterfactual scores. For a given concept, the candidate representation is constructed as the set of all voxels with positive specificity scores or a predefined number of voxels ranked the highest. This voxel set serves as the candidate region representing the concept. 
BrainTRACE can then evaluate the region as a whole by averaging activation and specificity scores across all voxels in the set. 
This way, the discovered region is required not only to respond strongly to positive images, but also to respond more strongly to them than to semantic negatives and counterfactual edits. The specificity score of the discovered region on the training set is used to determine whether it is a strong candidate representation, while final quantitative evaluation is carried out on a separate evaluation split and on measured fMRI data.

\subsection{Final Verdict and Follow-Up Experiment Design}
\label{sec:method:Steps:Final}

The final verdict in BrainTRACE is determined by two main quantities. The first is the \emph{\textbf{specificity evidence}} for the discovered candidate representation, measured by its specificity scores on the generated evaluation dataset and on the measured-data evaluation. High scores on both indicate stronger evidence that the voxel set responds selectively to the target concept rather than to correlated alternatives.
\emph{\textbf{Confidence Score}} is measured using statistical tests on our proposed scores, yielding for each concept representation a p-value for each score (see Appendix~\ref{sec_sup:statistical_test}). One can then choose the required p-value threshold and whether all significance tests should be passed, including activation and specificity scores on both generated and measured data.

The second is the \emph{\textbf{concept coverage in the measured data}}, estimated by analyzing what fraction of the desired positive and semantic-negative images can be successfully retrieved and verified in the measured image--fMRI dataset (see Appendix~\ref{sec_sup:retrieval_statistics}). This quantity reflects how informative the measured data is for validating the concept. For example, if only a few measured images contain dogs, then even a low or high measured-data evaluation score is less informative, since the dataset provides only limited evidence about that concept.
Based on the specificity evidence and the concept coverage in the measured data, the framework returns a final decision together with the candidate representation and, when relevant, a set of informative images for follow-up experiments. 
When concept coverage in the measured data is high, the measured-data evaluation is considered informative. In this case, strong specificity evidence leads to a high-confidence discovery, while weak specificity evidence leads to rejection. In contrast, when concept coverage in the measured data is low, the measured-data evaluation is less informative. In this case, strong specificity evidence on the generated dataset is treated as a promising but insufficiently validated finding and triggers follow-up experimental recommendations, whereas weak specificity evidence is treated as inconclusive.

\begin{figure}[t]
    \centering
    \includegraphics[width=1\textwidth]{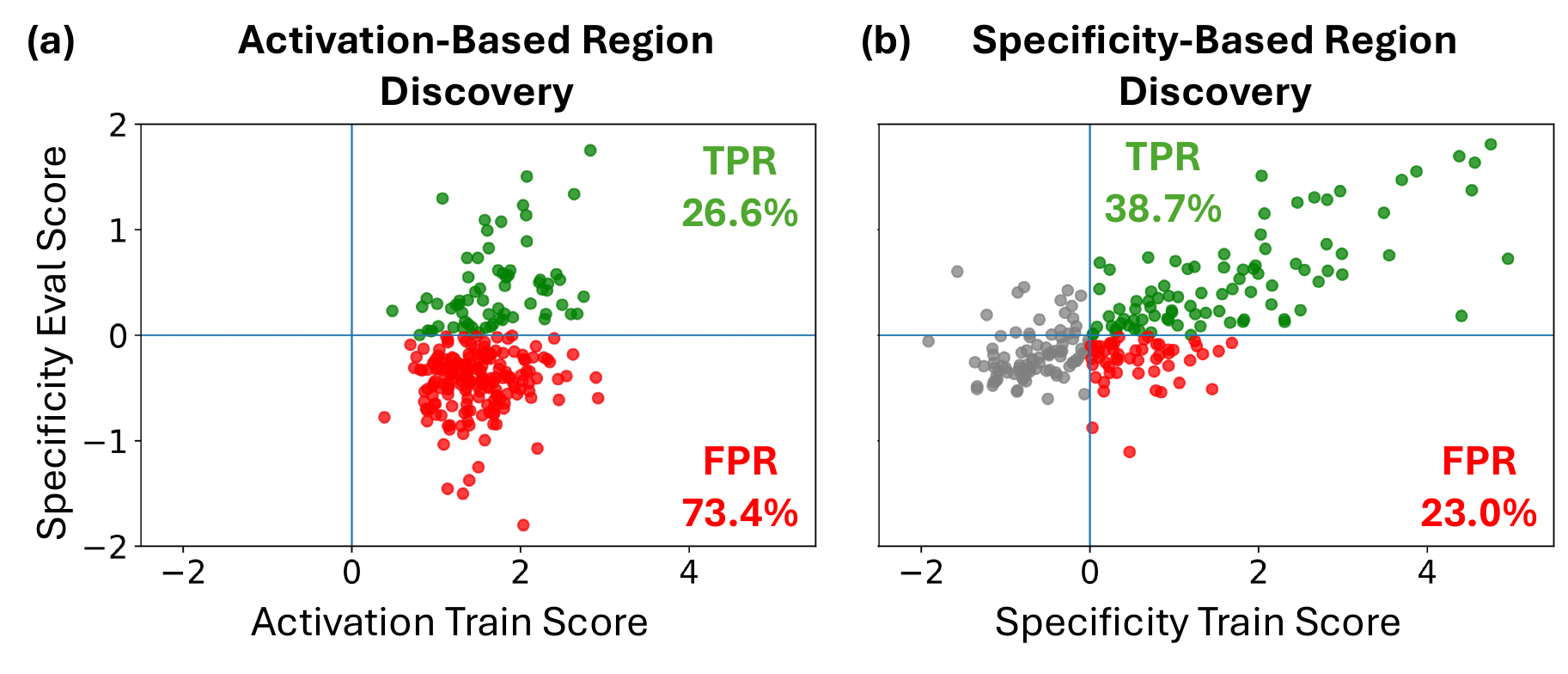}
    \caption{\textbf{Specificity-based ranking reduces false discoveries.}
    Activation-based discovery frequently selects regions that respond strongly to the target concept but do not exhibit specificity under the tested alternatives, leading to many false positives. By ranking candidates using specificity score, BrainTRACE suppresses correlation-driven discoveries, reduces false positives, and recovers more faithful concept representations.}
    \label{fig:Causal_vs_Corr}
\end{figure}

\section{Results}
\label{sec:Results}

BrainTRACE supports both counterfactual-specificity evaluation of candidate regions proposed by existing methods and discovery of new regions that more faithfully represent target concepts. We evaluate on the \emph{Natural Scenes Dataset} (NSD)~\cite{allen2022massive}, a large-scale 7-Tesla fMRI dataset consisting of 8 subjects each viewing approximately 10,000 natural images, reporting results on the 4 subjects who completed all sessions.

\subsection{BrainTRACE Counterfactual-specificity Testing}

Throughout this section we compare four region discovery methods: (i)~\textbf{Max Activation} ranks voxels by their average activation on generated positive images, following the standard localization approach. (ii)~\textbf{MindSimulator}~\cite{Bao2025MindSimulator} retrieves images from COCO using CLIP scores and ranks voxels by predicted fMRI response. For a fair comparison we use the same image--fMRI encoder as in BrainTRACE and the same number of overall tested images. We also compare (iii)~\textbf{MindSimulator+VLM}, which extends MindSimulator by using our filtering procedure with a vision-language model, to filter retrieved images that do not clearly depict the target concept. Finally, our method (iv)~\textbf{BrainTRACE} ranks voxels using a score combining both activation and specificity. For all methods, for the quantitative evaluation the discovered region per concept consists of the top 100 voxels ranked by the method score; results for other region sizes are provided in Appendix \ref{sec_sup:Region_size}.

\begin{figure}[t]
\centering

\begin{minipage}[t]{0.50\textwidth}
\centering
\small
\renewcommand{\arraystretch}{1.25}
\setlength{\tabcolsep}{3pt}

\captionof{table}{
\textbf{Quantitative comparison across discovery methods.}
Average activation and specificity scores for regions discovered for the top 50 concepts, averaged across four subjects. BrainTRACE preserves high activation while improving specificity.
}

\label{tab:quant_main}

\vspace{0.05cm}

\begin{tabular}{@{}lccccc@{}}
\toprule
& \multicolumn{2}{c}{\textbf{Activation}}
& \multicolumn{3}{c}{\textbf{Specificity}} \\
\cmidrule(lr){2-3}\cmidrule(lr){4-6}
\textbf{Method}
& \textbf{Gen.}
& \textbf{Meas.}
& \textbf{Gen.}
& \textbf{Meas.}
& \textbf{Edits} \\
\midrule
Max Activation
& \textbf{2.76} & 0.70 & 0.08 & 0.18 & 0.44 \\
MindSimulator
& 1.89 & 1.02 & -0.44 & 0.27 & 0.23 \\
MindSimulator+
& 2.13 & \textbf{1.12} & -0.26 & 0.41 & 0.38 \\
\emph{\textbf{BrainTRACE}}
& 2.05 & 1.08 & \textbf{0.62} & \textbf{0.71} & \textbf{0.98} \\
\bottomrule
\end{tabular}

\end{minipage}\hfill%
\begin{minipage}[t]{0.47\textwidth}
\centering
\small
\renewcommand{\arraystretch}{1.25}
\setlength{\tabcolsep}{10pt}

\captionof{table}{
\textbf{BrainTRACE aligns with known functional regions.}
Fraction of top-ranked voxels that fall within the expected functional regions for broad visual categories show high alignment with established functional organization.
}
\label{tab:known_roi_sanity}

\vspace{0.05cm}

\begin{tabular}{@{}lccc@{}}
\toprule
& \multicolumn{3}{c}{\textbf{Voxel Alignment Accuracy}} \\
\cmidrule(l){2-4}
\textbf{Region} & Top 100 & Top 200 & Top 500 \\
\midrule
Bodies & 99\% & 99\% & 97\% \\
Faces  & 90\% & 87\% & 84\% \\
Places & 74\% & 75\% & 74\% \\
Words  & 99\% & 98\% & 97\% \\
\bottomrule
\end{tabular}

\end{minipage}

\end{figure}

\vspace{-0.15cm}
\paragraph{BrainTRACE Evaluates Region Specificity.}
Fig~\ref{fig:Good_and_bad_Casulity_Edits} shows examples of our method's counterfactual-specificity evaluation on two regions supported by specificity evaluation (top) and two regions that fail specificity evaluation (bottom). For Social Scene and Text, not only is the positive activation high, but the semantic negatives and counterfactual edits show much lower activation. The counterfactual edits are particularly strong: for example, for Text, signs containing text are replaced by the same sign and background but with shapes or paint instead of text, while semantic negatives include visually similar but distinct concepts such as sketches. In the bottom rows, regions discovered by activation-based methods show high activation for positives but similarly high activation for negatives. Surfing, for example, was presented as a discovered representation in MindSimulator, but counterfactual-specificity testing reveals that the region discovered by this method likely responds to correlated factors such as water, human presence, or body pose rather than to surfing itself.

\vspace{-0.15cm}
\paragraph{Activation-Based Regions Have High False Positive Rate.}
Fig.~\ref{fig:Causal_vs_Corr}a shows, for each of 260 concepts and subject 1, the activation score and specificity score of the region discovered by MindSimulator approach. Each point represents one concept, with the activation score computed on the train set and the specificity score on the evaluation set. Concepts with a positive activation score are considered discovered by the method. We color in green those that are also supported by specificity evaluation (true positives, TPR) and in red those with high activation but negative specificity score (false positives, FPR). Activation-based discovery yields a false positive rate of nearly 70\%, meaning that most discovered regions are in fact driven by correlated factors rather than the target concept.

\vspace{-0.15cm}
\paragraph{Specificity-Based Ranking Discovers More Faithful Regions.}
Fig.~\ref{fig:Causal_vs_Corr}b shows the same plot but with regions chosen according to the train specificity score. Concepts with a negative train specificity score are labeled as not supported by specificity evaluation and withheld from discovery (grey points), avoiding false positives. This reduces the false positive rate from 73.4\% to 23\%, while improving the true positive rate from 26.6\% to 38.7\%. Together, this shows that specificity-based ranking not only avoids spurious localizations but also recovers more genuinely concept-specific regions than activation-based methods.

\vspace{-0.15cm}
\paragraph{Regions Identified by Specificity-Based Ranking Maintain High Activation While Improving Specificity Scores.}
Table~\ref{tab:quant_main} compares all four methods across activation and specificity scores. For each method we select the top 50 concepts by train score and report results on regions of 100 voxels (other region sizes in the appendix). Activation is measured as the average response to positive images, either generated by our pipeline and held-out (Gen.) or retrieved from the held-out measured image--fMRI pool (Meas.). Specificity is measured via semantic negative scores (on both measured and generated data) and counterfactual edit scores. Activation-based methods achieve high activation scores but show notably low semantic negative scores, with other methods scoring below zero, indicating non-concept-specific regions. BrainTRACE achieves comparable activation scores while substantially improving specificity: for example, the semantic negative score on generated data improves from -0.44 (MindSimulator) to 0.62 (BrainTRACE), and on measured data from 0.27 to 0.71. This confirms that BrainTRACE discovers regions that are both strongly activated by the target concept and specific to it.

\begin{figure}[t]
    \centering
    \includegraphics[width=1.0\textwidth]{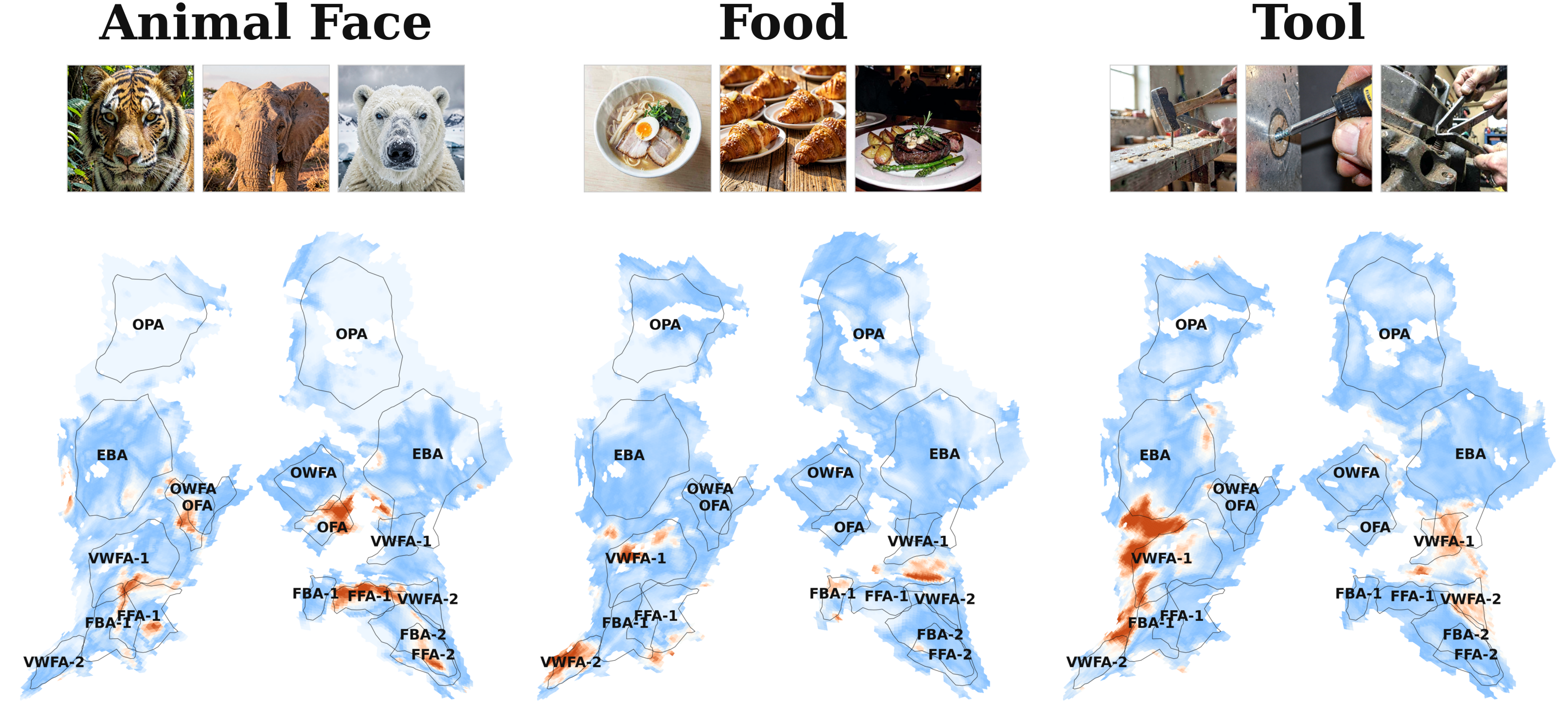}
    
    \caption{\textbf{Concepts discovered by counterfactual-specificity evaluation.} 
    We show voxel-wise specificity scores on brain maps for three example concepts, with representative positive images above each map. Each panel shows a flatmap of high-level visual cortex for subject 1. Each voxel is colored by its specificity score, where warmer colors indicate higher concept-specific specificity evidence. Black outlines and labels mark NSD functional ROIs, allowing comparison with known visual regions.}
    \label{fig:brain_maps}
    \vspace{-0.15cm}
\end{figure}

\subsection{Fine-Grained Visual Concept Localization}

\paragraph{Alignment with Known Functional Regions.}
To validate our method, we first examine whether regions identified through counterfactual-specificity testing align with well-established functional areas, focusing on four broadly studied categories: Bodies, Faces, Places, and Words. For each category, we select the relevant concepts from our set, rank voxels by specificity score, and measure the fraction of the top \(K\) voxels that fall within the corresponding NSD-localized functional region. Table~\ref{tab:known_roi_sanity} shows results for voxel set sizes of 100, 200, and 500, demonstrating strong alignment across all four categories.

\paragraph{Discovery of Fine-Grained Concept Representations.}
Fig.~\ref{fig:brain_maps} shows flatmaps of high-level visual cortex in which each voxel is colored by its specificity score, from low (blue) to high (red). Despite imposing no spatial constraints, the discovered regions are spatially localized, whereas activation-based methods produce many more high-scoring voxels (see Appendix~\ref{sec_sup:activation_vs_ours_maps}), reflecting the greater selectivity of specificity-based scoring. As examples, the concept \emph{tools} localizes near body-part and action-related regions such as EBA, while \emph{animal faces} localizes within known face-selective areas such as FFA and OFA. More broadly, BrainTRACE discovers localized representations supported by specificity testing for many different concepts (see Appendix for additional examples). Fig.~\ref{fig:Clusters_and_brains} further shows binary maps of voxels with positive specificity scores for sets of related concepts, revealing fine-grained distinctions between them: body-part concepts such as \emph{human face}, \emph{human hands}, and \emph{human legs} divide activity across face areas (FFA) and body-related regions (EBA and FBA), while text-related concepts such as \emph{handwritten text}, \emph{symbolic signs}, and \emph{logos} show distinct patterns within VWFA- and OWFA-related regions.

\paragraph{Consistency Across Subjects.}
Appendix~\ref{sec_sup:subject_alignment} shows brain maps for the same concepts across subjects. Despite the well-known variability of functional organization across individuals, we observe clear correspondence in the region locations where concepts are represented. This both strengthens the validity of our findings and suggests that the discovered concept-specific representations capture robust aspects of visual cortex organization.

\begin{figure}[h]
    \centering
    \includegraphics[width=1.0\textwidth]{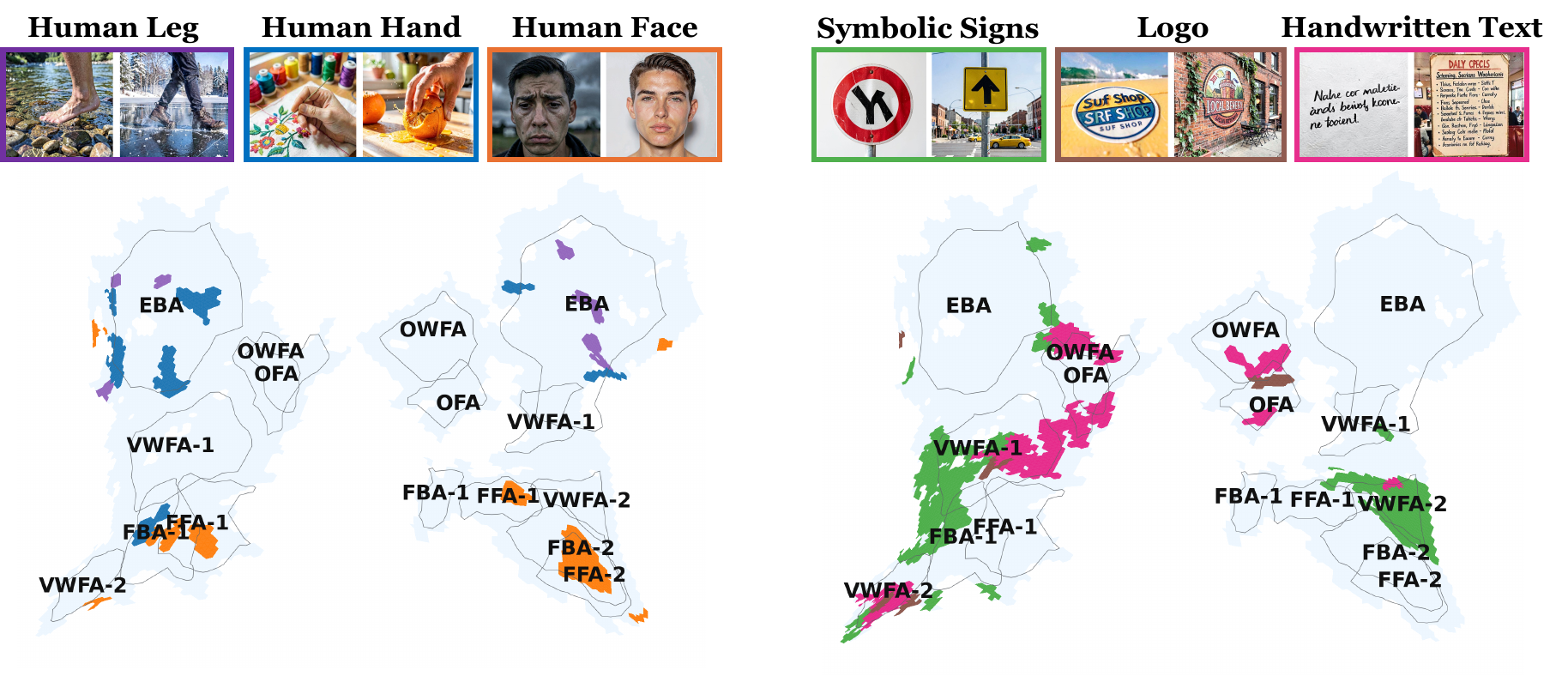}
    \caption{\textbf{Fine-grained organization of related concepts.} Left: body-related concepts, including human face, human hand, and human leg, show distinct voxel patterns across face- and body-selective regions. Right: text-related concepts, including handwritten text, symbolic signs, and logos, show distinct voxel patterns across word- and object-related visual areas. These results show that BrainTRACE discovers nearby semantic categories within high-level visual cortex.}
    \label{fig:Clusters_and_brains}
\end{figure}

\subsection{Ablation and Analysis}

In Appendix~\ref{sec_sup:Ablation}, we provide a detailed analysis of multiple factors and results. We examine the contribution of different ranking scores, including both activation-based and specificity scores (Appendix~\ref{sec_sup:ranking_score_analysis}), as well as the effect of the number of voxels chosen to define a concept region, showing that scores gradually decrease as region size increases (Appendix~\ref{sec_sup:Region_size}). We also examine quantitative results across subjects, comparing all four analyzed NSD subjects and showing that the advantage of specificity-based ranking remains (Appendix~\ref{sec_sup:Quantitative_across_sbuecjts}), as well as qualitative results showing consistency of the discovered representations across subjects (Appendix~\ref{sec_sup:subject_alignment}). We further analyze the coverage of retrieved positives and semantic negatives for each concept in the measured dataset, both to quantify the reliability of each concept’s measured-data evaluation and to identify missing negative concepts that may be important for future experiments (Appendix~\ref{sec_sup:retrieval_statistics}). Finally, we analyze some of the false positive cases of BrainTRACE. Although substantially fewer than in activation-based methods, these cases still occur and are often associated with broad image properties, such as \emph{sky}, \emph{reflection}, and \emph{lighting contrast}. By inspecting their highest-activating negatives, we find many cases of semantic-negative generation failures, reflecting current limitations of the underlying language and vision models in generating semantically similar concepts while excluding the target concept itself (Appendix~\ref{sec_sup:negative_generation_fail}).

\vspace{-0.1cm}
\section{Conclusions}
\label{sec:Conclusions}
\vspace{-0.1cm}

A central message of this work is the importance of testing concept specificity in visual representation discovery in the brain. While the distinction between responses driven by correlated factors and truly concept-specific representations has been widely discussed in other fields, and in some areas of neuroscience, it has received more limited attention in the study of visual concept representations from fMRI.
BrainTRACE takes a step toward addressing this gap by supporting counterfactual-specificity evaluation and enabling discovery of more specific and reliable concept representations, including fine-grained concepts beyond broad category-level regions. At the same time, BrainTRACE depends on current language and vision models and is therefore not free of failure modes. Errors in image generation, editing, retrieval, verification, or in the scope of the proposed counterfactuals may still leave relevant correlated factors untested, meaning that some discovered representations may remain specific only with respect to the alternatives considered by the current pipeline. We expect this limitation to improve as language and vision models continue to advance. An important direction for future work is to make the pipeline more iterative, allowing current results and activation patterns to guide the proposal of new counterfactuals and semantic negatives for stronger validation. More broadly, BrainTRACE, together with its analysis of which concepts and negative controls are sufficiently represented in measured data, provides a step toward counterfactual-specificity testing of brain representations and toward tighter integration between computational and experimental neuroscience.

\section*{Acknowledgments}
This research was partially supported by the European Research Council (ERC) under the Horizon programme, grant number 101142115. We are also grateful for the support of ARL, MIT-IBM Watson AI Lab, Hyundai Motor Company and ONR MURI.

\bibliographystyle{unsrtnat}
\bibliography{references}

\newpage
\appendix
\begin{center}
    {\LARGE \bfseries Appendix}
\end{center}
\renewcommand{\thefigure}{S\arabic{figure}}  
\renewcommand{\thetable}{T\arabic{table}}    

\setcounter{figure}{0} 
\setcounter{table}{0}

\section{Ablation \& Analysis}
\label{sec_sup:Ablation}

\subsection{Consistency Across Subjects}
\label{sec_sup:subject_alignment}

\begin{figure}[h]
    \centering
    \includegraphics[width=1.0\textwidth]{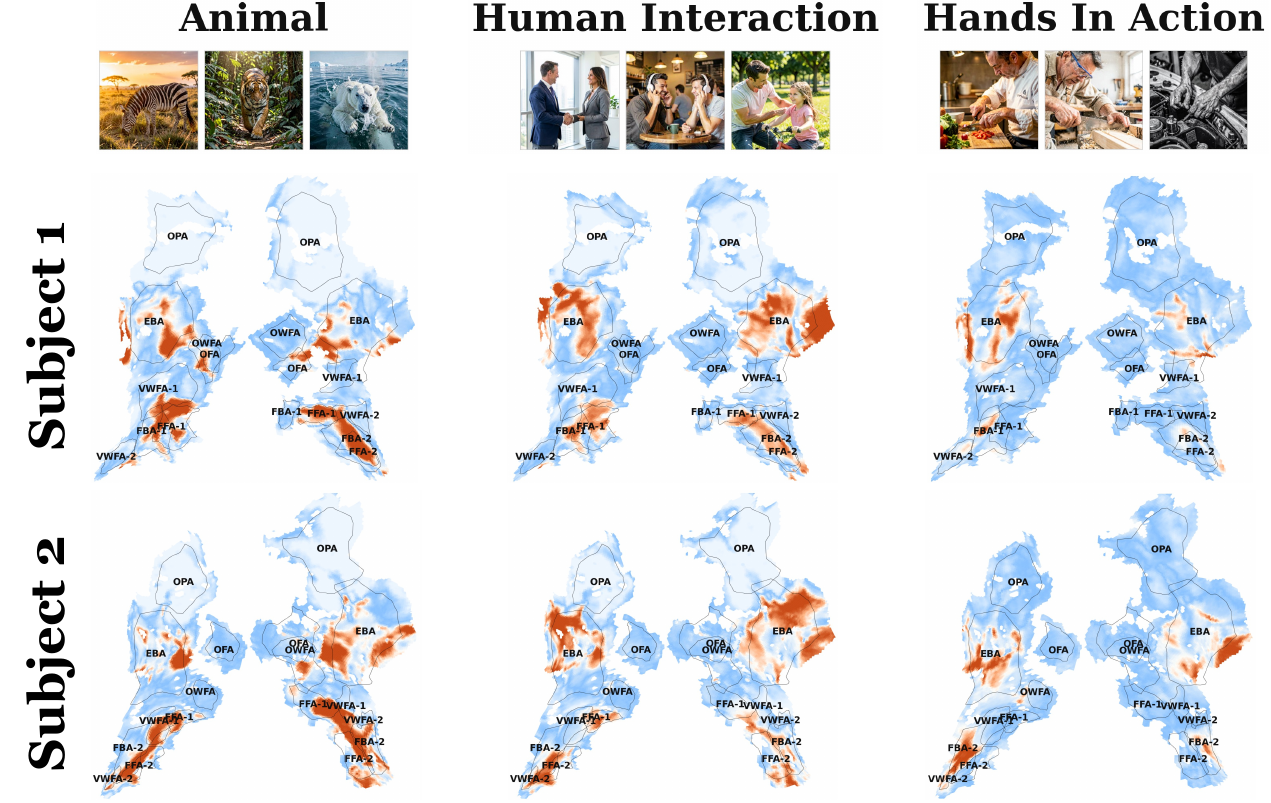}
    \caption{\textbf{Cross-subject consistency of concept representations identified through specificity testing.} Cortical flatmaps show BrainTRACE specificity scores for three representative concepts, Animal, Human Interaction and Hands in Action, across two NSD subjects. Example positive images for each concept are shown above the corresponding maps. Warmer colors indicate higher concept-specific specificity evidence, while cooler colors indicate lower scores. Across subjects, high-scoring voxels appear in similar high-level visual regions, demonstrating that BrainTRACE discovers spatially localized and reproducible representations despite individual variability in cortical organization.}
    \label{fig:Two_subjects_maps}
\end{figure}

While different subjects show substantial variability, even in their responses to the same image, observing correspondence in the localization of discovered representations across subjects provides strong validation of the method. It is also of independent interest, as it suggests that aspects of visual organization are shared across subjects despite individual variability. To examine this, we compare voxel-wise specificity maps for three concepts—\emph{Animal}, \emph{Human Interaction} and \emph{Hands in Action} —across Subjects 1 and 2. Both the cortical flatmaps and the marked functional regions, taken from NSD, differ across subjects, and the specificity maps themselves are not identical. Nevertheless, as shown in Fig.~\ref{fig:Two_subjects_maps}, for all three concepts the main representation regions show clear correspondence across subjects. This further supports the validity of the discovered regions and suggests that visual concept representations identified through counterfactual-specificity testing capture shared aspects of cortical organization.

\clearpage
\subsection{Comparison of Specificity and Activation Maps}
\label{sec_sup:activation_vs_ours_maps}

\begin{figure}[h]
    \centering
    \includegraphics[width=1.0\textwidth]{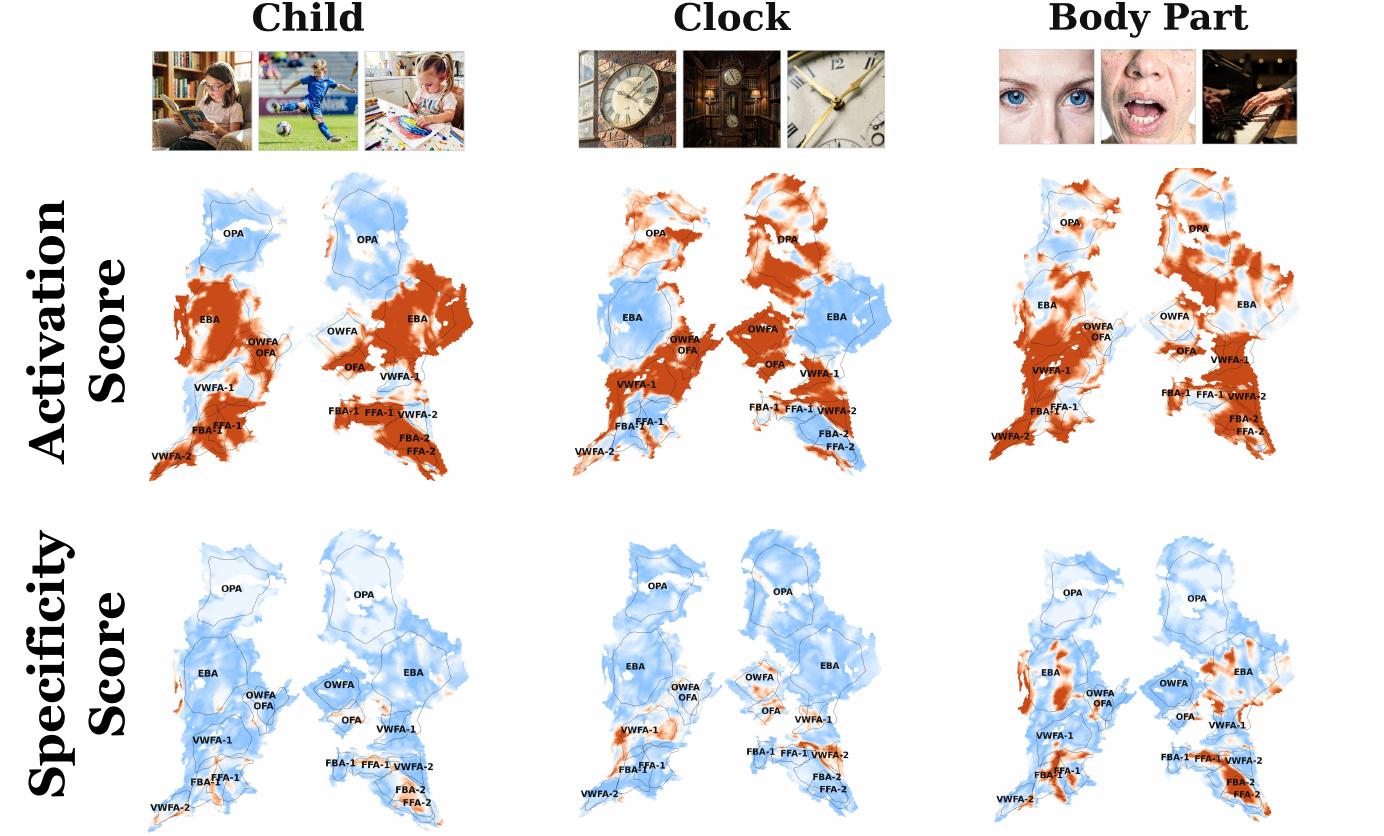}
    \caption{\textbf{Specificity versus activation-based localization across diverse concepts.} Maps compare BrainTRACE specificity scoring with activation-based localization for Child, Clock, and Body Part. Across these examples, activation-based localization produces broad high-response patterns, while specificity scoring yields more selective maps by suppressing responses driven by correlated visual or semantic cues. }
    \label{fig:activation_based_vs_ours_maps}
\end{figure}

We compare the regions discovered by BrainTRACE to those obtained by a standard activation-based localization method, in order to understand how specificity scoring changes the resulting concept maps. The main difference is that activation-based localization tends to assign high scores to broad sets of voxels, often producing large regions that likely reflect correlated visual or semantic factors rather than concept-specific representations. In contrast, BrainTRACE uses specificity scoring to favor voxels whose responses remain stronger for the target concept than for both semantic negatives and counterfactual edits, leading to much more selective candidate regions. This increased selectivity is important for two reasons: it reduces the false positive rate of concept discovery, and it also allows the method to conclude that some concepts are not reliably represented rather than forcing a broad activation pattern into a putative localization. Fig.~\ref{fig:activation_based_vs_ours_maps} illustrates this contrast for several example concepts. Across all of them, the activation-based maps are substantially broader; for example, for \emph{Child}, almost the entire EBA receives high activation scores, which is unlikely to reflect a representation specific to that concept alone. In contrast, the BrainTRACE maps isolate smaller and more plausible concept-specific regions.

\clearpage

\subsection{Analysis of Ranking Scores}
\label{sec_sup:ranking_score_analysis}

Table~\ref{tab_sup:Abaltion_scores} analyzes how different ranking criteria affect the quality of the discovered regions. We analyze the following ranking criteria: Max Activation–Generated (MAG), Max Activation–Retrieved from Measured (MAM), Max Activation–Retrieved from Pool (MAL), Max Activation–Retrieved from Filtered Pool (MALF), Semantic Specificity-Generated (SSG), Semantic Specificity-Retrieved from Measured (SSM), Semantic Specificity-Retrieved from Pool (SSL), and Counterfactual Specificity-Generated (CFG).

Several trends emerge. First, ranking by \textbf{max activation} alone leads to the highest activation scores, especially when positives are generated (MAG) or retrieved from measured data (MAM), but performs substantially worse on the specificity criteria, particularly on generated semantic negatives. This indicates that activation-based ranking often identifies regions that respond strongly to the target concept images, yet are less selective when challenged with correlated alternatives. Retrieval from measured data improves measured activation and measured specificity scores relative to retrieval from the large image pool, while filtering the retrieved pool also improves specificity performance, suggesting that retrieval quality has a strong effect on the final region quality.

Second, ranking by the specificity criteria alone shifts the balance. \textbf{Semantic Specificity – Generated (SSG)} achieves the strongest generated specificity score, while \textbf{Semantic Specificity – Retrieved from Measured (SSM)} gives the strongest measured specificity score. \textbf{Counterfactual Specificity – Generated (CFG)} achieves by far the best edit score. These results support the intended role of each ranking signal: generated semantic negatives best enforce discrimination against correlated generated alternatives, measured retrieval strengthens performance on measured-data evaluation, and edit-based ranking most strongly captures sensitivity under controlled counterfactual edits.

Finally, combining the different signals provides the best overall balance. Adding semantic-specificity and edit-based scores already improves the average score over the single-score variants, and further adding activation-based signals increases activation performance without collapsing the specificity scores. The best overall result is obtained by \textbf{CFG+SSG+SSL+MALF+SSM}, which achieves the highest average score across all criteria. This suggests that no single ranking signal is sufficient on its own: strong region discovery requires combining activation, generated semantic negatives, measured semantic negatives, and edit-based counterfactual-specificity testing. Overall, the table supports the main design choice of BrainTRACE, namely that the most reliable regions are obtained by integrating multiple complementary signals rather than relying on activation alone.

\begin{table}[ht]
\centering
\setlength{\tabcolsep}{3.6pt}
\caption{\textbf{Analysis of different ranking scores.} Scores are reported for the 50 top-ranked concepts, averaged across subjects, using regions of size 100. Rows compare different ranking criteria used to define the voxel region. Max Activation ranks voxels by their average activation across images, using positives generated by the model, retrieved from measured images, or retrieved from a large image pool, with and without filtering. Semantic Specificity subtracts the strongest negative response, using the top-10 negative images. Counterfactual Specificity compares each positive image to its corresponding edited negatives by subtracting the maximum edited response. Combination rows aggregate these signals; the final combination provides the best balance across activation and specificity criteria.}
\vspace{+0.12cm}
\begin{tabular}{@{}lcccccc@{}}
\toprule
& \multicolumn{2}{c}{\textbf{Activation}}
& \multicolumn{3}{c}{\textbf{Specificity}}
&  \\
\cmidrule(lr){2-3}\cmidrule(lr){4-6}
\textbf{Method}
& \textbf{Gen.}
& \textbf{Meas.}
& \textbf{Gen.}
& \textbf{Meas.}
& \textbf{Edits}
& \textbf{Avg.} \\
\midrule

Max Activation - Generated (MAG)
& \textbf{2.76} & 0.70 & 0.08 & 0.18 & 0.44 & 0.83 \\

Max Activation - Retrieved from Measured (MAM)
& 1.71 & \textbf{1.83} & -0.33 & 0.54 & 0.22 & 0.79 \\

Max Activation - Retrieved from Pool (MAL)
& 1.89 & 1.02 & -0.44 & 0.27 & 0.23 & 0.59 \\

Max Activation - Retrieved from Filtered Pool (MALF)
& 2.13 & 1.12 & -0.26 & 0.41 & 0.38 & 0.76 \\

Semantic Specificity - Generated (SSG)
& 1.73 & 0.37 & \textbf{0.94} & 0.09 & 0.71 & 0.77 \\

Semantic Specificity - Retrieved from Measured (SSM)
& 1.29 & 1.18 & -0.10 & \textbf{0.99} & 0.39 & 0.75 \\

Semantic Specificity - Retrieved from Pool (SSL)
& 1.43 & 0.74 & 0.16 & 0.50 & 0.63 & 0.69 \\

 Counterfactual Specificity - Generated (CFG)
& 1.67 & 0.56 & 0.45 & 0.21 & \textbf{1.42} & 0.86 \\

CFG+SSG+SSL
& 1.90 & 0.65 & 0.82 & 0.37 & 1.12 & 0.97 \\

CFG+SSG+SSL+MAG
& 2.29 & 0.73 & 0.81 & 0.39 & 1.08 & 1.06 \\

CFG+SSG+SSL+MALF
& 2.10 & 0.84 & 0.70 & 0.45 & 1.05 & 1.03 \\

CFG+SSG+SSL+MALF+SSM
& 2.05 & 1.08 & 0.62 & 0.71 & 0.98 & \textbf{1.09} \\

\bottomrule
\end{tabular}
\label{tab_sup:Abaltion_scores}
\end{table}

\clearpage

\subsection{Region Size Analysis}
\label{sec_sup:Region_size}

Table~\ref{tab:scores_top50_all_roi_sizes} analyzes how the quality of the discovered regions changes as the region size increases from 50 to 1000 voxels. A clear pattern emerges across all sizes. Activation-based methods, especially \textbf{Max Activation}, consistently achieve the strongest activation scores on generated positives, while \textbf{MindSimulator+} often achieves the highest measured activation scores. However, these gains come at the cost of substantially weaker specificity scores. In contrast, \emph{\textbf{BrainTRACE}} consistently obtains the strongest specificity scores across all region sizes and across all three counterfactual-specificity evaluation criteria: generated negatives, measured negatives, and counterfactual edits. As the region size increases, scores gradually decrease for all methods, indicating that larger regions become less selective. Nevertheless, the relative advantage of \emph{\textbf{BrainTRACE}} remains stable, showing that the proposed specificity ranking yields regions that are more faithful to the target concept across a broad range of region sizes.

\begin{table}[h]
\centering
\setlength{\tabcolsep}{7pt}
\caption{\textbf{Scores across region sizes for the top 50 discovered concepts.}
Average activation and specificity
validation scores for regions discovered by each method, evaluated at different region sizes. For each region size, scores are averaged over the top 50 concepts selected by the corresponding method. Max Activation ranks voxels by average activation on generated positives; MindSimulator ranks voxels by predicted responses to CLIP-retrieved concept images; MindSimulator+ additionally filters retrieved images using VLM verification; and BrainTRACE ranks voxels using the combined specificity score. As region size increases, scores gradually decrease for all methods, reflecting the reduced selectivity of larger regions. Nevertheless, BrainTRACE maintains relatively high activation while consistently achieving substantially stronger specificity scores on generated negatives, measured negatives, and counterfactual edits across all tested region sizes.}
\vspace{+0.17cm}
\begin{tabular}{@{}clccccc@{}}
\toprule
& 
& \multicolumn{2}{c}{\textbf{Activation}}
& \multicolumn{3}{c}{\textbf{Specificity}} \\
\cmidrule(lr){3-4}\cmidrule(lr){5-7}
\makecell{\textbf{ROI}\\\textbf{Size}}
& \textbf{Method}
& \textbf{Gen.}
& \textbf{Meas.}
& \textbf{Gen.}
& \textbf{Meas.}
& \textbf{Edits} \\
\midrule

\multirow{4}{*}{50}
& Max Activation
& \textbf{2.91} & 0.65 & 0.08 & 0.16 & 0.46 \\
& MindSimulator
& 1.97 & 1.00 & -0.45 & 0.26 & 0.24 \\
& MindSimulator+
& 2.22 & 1.07 & -0.26 & 0.39 & 0.40 \\
& \emph{\textbf{BrainTRACE}}
& 2.13 & \textbf{1.08} & \textbf{0.68} & \textbf{0.71} & \textbf{1.06} \\

\addlinespace[2pt]
\midrule

\multirow{4}{*}{100}
& Max Activation
& \textbf{2.76} & 0.70 & 0.08 & 0.18 & 0.44 \\
& MindSimulator
& 1.89 & 1.02 & -0.44 & 0.27 & 0.23 \\
& MindSimulator+
& 2.13 & \textbf{1.12} & -0.26 & 0.41 & 0.38 \\
& \emph{\textbf{BrainTRACE}}
& 2.05 & 1.08 & \textbf{0.62} & \textbf{0.71} & \textbf{0.98} \\

\addlinespace[2pt]
\midrule

\multirow{4}{*}{200}
& Max Activation
& \textbf{2.61} & 0.74 & 0.06 & 0.19 & 0.40 \\
& MindSimulator
& 1.78 & 1.01 & -0.47 & 0.25 & 0.19 \\
& MindSimulator+
& 2.01 & \textbf{1.12} & -0.30 & 0.39 & 0.32 \\
& \emph{\textbf{BrainTRACE}}
& 1.95 & 1.04 & \textbf{0.55} & \textbf{0.69} & \textbf{0.90} \\

\addlinespace[2pt]
\midrule

\multirow{4}{*}{500}
& Max Activation
& \textbf{2.35} & 0.72 & 0.01 & 0.17 & 0.35 \\
& MindSimulator
& 1.60 & 0.96 & -0.49 & 0.21 & 0.14 \\
& MindSimulator+
& 1.79 & \textbf{1.05} & -0.32 & 0.34 & 0.28 \\
& \emph{\textbf{BrainTRACE}}
& 1.77 & 0.96 & \textbf{0.45} & \textbf{0.62} & \textbf{0.80} \\

\addlinespace[2pt]
\midrule

\multirow{4}{*}{1000}
& Max Activation
& \textbf{2.12} & 0.64 & 0.00 & 0.16 & 0.31 \\
& MindSimulator
& 1.42 & 0.90 & -0.52 & 0.16 & 0.09 \\
& MindSimulator+
& 1.62 & \textbf{0.98} & -0.36 & 0.27 & 0.21 \\
& \emph{\textbf{BrainTRACE}}
& 1.59 & 0.87 & \textbf{0.36} & \textbf{0.55} & \textbf{0.70} \\

\bottomrule
\end{tabular}
\label{tab:scores_top50_all_roi_sizes}
\vspace{-0.3cm}
\end{table}

\clearpage
\subsection{Quantitative Results Across Subjects}
\label{sec_sup:Quantitative_across_sbuecjts}

Table~\ref{tab:scores_top50_across_subjects_region100} compares the quantitative results across the four NSD subjects used in our analysis (1, 2, 5, and 7), who completed all sessions, using regions of size 100 and the top 50 concepts selected by each method. The same overall pattern is consistent across subjects. Activation-based methods, especially Max Activation, achieve the strongest generated activation scores, while MindSimulator+ often gives the highest measured activation scores. However, BrainTRACE consistently yields the strongest specificity scores on generated negatives, measured negatives, and counterfactual edits for every subject. This suggests that the benefit of specificity-based ranking is not specific to a single subject, but generalizes across individuals.

\begin{table}[ht]
\centering
\setlength{\tabcolsep}{4pt}
\caption{\textbf{Scores on the top 50 concepts across subjects (region size 100).}
Average activation and specificity validation scores for regions of size 100 discovered separately for each subject. 
For every subject, scores are averaged over the top 50 concepts selected by the corresponding method. 
The four subjects shown are the NSD subjects who completed all scanning sessions. 
Max Activation ranks voxels by average activation on generated positives; MindSimulator ranks voxels by predicted responses to CLIP-retrieved concept images; MindSimulator+ additionally filters retrieved images using VLM verification; and BrainTRACE ranks voxels using the combined specificity score. 
Across subjects, activation-based methods achieve the highest activation scores, while BrainTRACE consistently attains stronger specificity scores on generated negatives, measured negatives, and counterfactual edits.}
\vspace{+0.92cm}
\begin{tabular}{@{}clccccc@{}}
\toprule
& 
& \multicolumn{2}{c}{\textbf{Activation}}
& \multicolumn{3}{c}{\textbf{Specificity}} \\
\cmidrule(lr){3-4}\cmidrule(lr){5-7}
\textbf{Subject}
& \textbf{Method}
& \textbf{Gen.}
& \textbf{Meas.}
& \textbf{Gen.}
& \textbf{Meas.}
& \textbf{Edits} \\
\midrule

\multirow{4}{*}{1}
& Max Activation
& \textbf{2.83} & 0.74 & 0.10 & 0.19 & 0.43 \\
& MindSimulator
& 1.96 & 1.04 & -0.49 & 0.30 & 0.26 \\
& MindSimulator+
& 2.20 & \textbf{1.19} & -0.21 & 0.46 & 0.43 \\
& \emph{\textbf{BrainTRACE}}
& 2.09 & 1.16 & \textbf{0.62} & \textbf{0.79} & \textbf{0.97} \\

\addlinespace[2pt]
\midrule

\multirow{4}{*}{2}
& Max Activation
& \textbf{2.84} & 0.77 & 0.04 & 0.15 & 0.45 \\
& MindSimulator
& 1.89 & 0.97 & -0.51 & 0.18 & 0.21 \\
& MindSimulator+
& 2.22 & \textbf{1.06} & -0.27 & 0.28 & 0.38 \\
& \emph{\textbf{BrainTRACE}}
& 2.17 & 0.99 & \textbf{0.65} & \textbf{0.61} & \textbf{0.99} \\

\addlinespace[2pt]
\midrule

\multirow{4}{*}{5}
& Max Activation
& \textbf{2.72} & 0.80 & 0.08 & 0.26 & 0.48 \\
& MindSimulator
& 1.95 & 1.24 & -0.39 & 0.39 & 0.23 \\
& MindSimulator+
& 2.12 & \textbf{1.35} & -0.34 & 0.55 & 0.39 \\
& \emph{\textbf{BrainTRACE}}
& 2.04 & 1.25 & \textbf{0.58} & \textbf{0.84} & \textbf{1.00} \\

\addlinespace[2pt]
\midrule

\multirow{4}{*}{7}
& Max Activation
& \textbf{2.67} & 0.50 & 0.11 & 0.13 & 0.39 \\
& MindSimulator
& 1.76 & 0.83 & -0.37 & 0.21 & 0.21 \\
& MindSimulator+
& 1.97 & \textbf{0.90} & -0.22 & 0.35 & 0.31 \\
& \emph{\textbf{BrainTRACE}}
& 1.90 & 0.89 & \textbf{0.61} & \textbf{0.62} & \textbf{0.96} \\

\bottomrule
\end{tabular}
\label{tab:scores_top50_across_subjects_region100}
\end{table}

\subsection{Retrieval Statistics and Missing Concept Coverage}
\label{sec_sup:retrieval_statistics}

A key part of BrainTRACE is understanding not only which concepts can be validated in the measured data, but also which concepts are \emph{not} sufficiently represented there. This is important for two reasons. First, low measured-data scores should be interpreted differently when the target concept itself is rarely present in the measured dataset. In such cases, weak measured validation may reflect lack of coverage rather than lack of a true brain representation. Second, this analysis helps guide future experiments by identifying which concepts and which negative controls are missing from the current dataset, and therefore which additional stimuli would be most informative to collect.

\begin{figure}[t]
    \centering
    \includegraphics[width=0.9\textwidth]{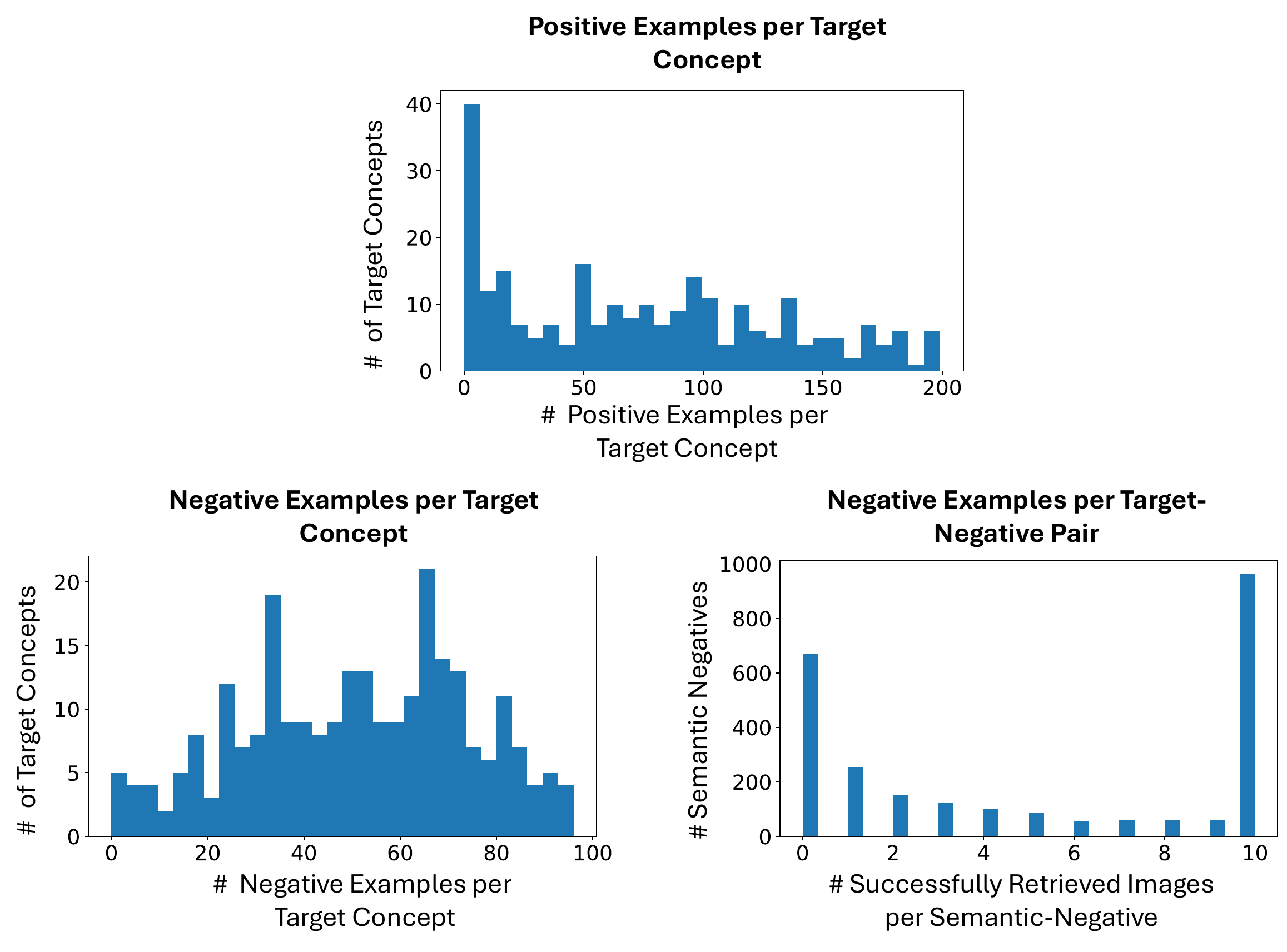}
    \vspace{-0.20cm}
    \caption{\textbf{Coverage of retrieved positives and semantic negatives in the measured dataset.}
    For each target concept, we count the number of successfully retrieved and verified positive examples (top) and semantic-negative examples (bottom left). We also report the number of successfully retrieved negatives for each target--negative pair (bottom right). This analysis is used to assess how informative the measured data is for validating discovered representations and for identifying missing evidence that may require follow-up experiments.}
    \label{fig:retrieval_statistics}
    \vspace{-0.6cm}
\end{figure}

We begin by analyzing the coverage of \emph{positive} examples in the measured image--fMRI dataset. As shown in the histogram of positive examples per target concept (Fig.~\ref{fig:retrieval_statistics}), the number of successfully retrieved and verified positive images varies substantially across concepts, ranging from very few examples to nearly 200. A considerable fraction of concepts have only a small number of measured positives, while others are much better represented. This variability is critical for interpreting measured-data validation: when a concept has only limited coverage, even strong or weak measured-data scores are less informative, since the available dataset provides only sparse evidence about that concept. In contrast, when many verified positive examples are available, measured-data validation can provide much stronger support for or against a discovered representation.
This retrieval-coverage analysis therefore provides an estimate of how informative the measured dataset is for each concept. It is used both to interpret the reliability of discovered representations and to identify concepts for which additional data would be especially valuable. In practice, concepts with high specificity score but low positive coverage are treated as less well validated by measured-data evaluation alone, and are natural candidates for follow-up experiments with targeted stimuli.

In addition to positive coverage, BrainTRACE also analyzes retrieval of \emph{semantic negatives}, which are important for validating specificity under counterfactual and semantic-negative comparisons on measured data. For each target concept, the language model proposes relevant negative concepts, and the retrieval pipeline searches for measured examples using CLIP-based scoring together with a double verification process: the target concept must be absent, and the intended negative concept must be present. The retrieval statistics further show substantial variability in the number of successfully retrieved semantic negatives, both per target concept and per target--negative pair. In particular, many target--negative pairs either yield no measured examples or only a few, while others reach the maximum retrieval count. This means that for some concepts, measured-data counterfactual evaluation is limited not only by missing positives but also by missing semantically informative negatives. Such cases are especially important for experimental planning, since they indicate which negative controls should be added in future fMRI studies to better distinguish true concept-specific responses from correlated alternatives.

\clearpage

\subsection{Analyzing BrainTRACE False Positives}
\label{sec_sup:negative_generation_fail}

\begin{figure}[h]
    \centering
    \includegraphics[width=1.0\textwidth]{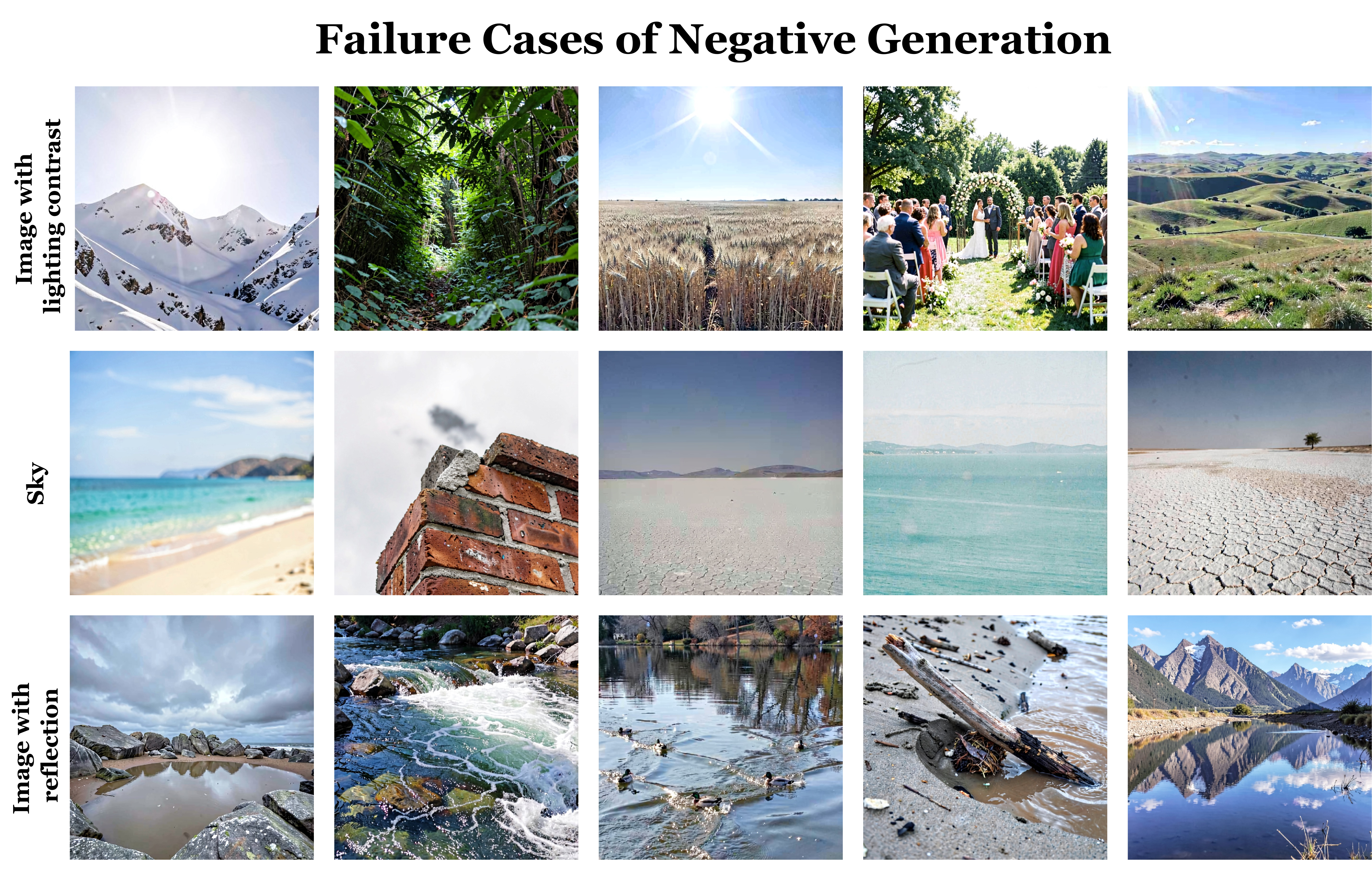}
\caption{\textbf{Failure cases in semantic-negative generation.} Representative examples for the target concepts \emph{sky}, \emph{reflection}, and \emph{lighting contrast}, where the generated semantic negatives still contain clear instances of the target concept.}
    \label{fig:negative_fail}
\end{figure}

Although BrainTRACE yields a much lower number of false positives than activation-based methods, some false positives still remain. We analyzed these cases and found that many involve broad, generic image properties rather than more localized visual concepts. Representative examples include \emph{lighting contrast}, \emph{sky}, and \emph{reflection}. In addition, while examining the pipeline outputs, we observed that a substantial fraction of the remaining failures are associated with semantic-negative generation failures.
The semantic-negative generation pipeline is designed to produce images that exclude the target concept, using both LLM-guided prompt construction and VLM-based verification. In practice, however, failures occasionally occur. Fig.~\ref{fig:negative_fail} shows representative examples for the target concepts \emph{sky}, \emph{reflection}, and \emph{lighting contrast}, where the generated negatives still visibly contain the target concept. For example, images generated for counter concepts related to outdoor scenes may still contain large sky regions, while negatives intended to exclude reflections may nevertheless include reflective surfaces such as water or glass. These cases highlight an important limitation of semantic-negative generation for broad visual properties that are difficult to isolate cleanly. This challenge is especially pronounced when the goal is to generate semantically similar concepts that exclude the target concept itself. This remains a limitation of the current pipeline, and we expect future advances in language and vision models to improve it.

\clearpage

\clearpage

\subsection{Statistical Testing of BrainTRACE Discovered Regions}
\label{sec_sup:statistical_test}

\begin{figure}[h]
    \centering
    \includegraphics[width=1.0\textwidth]{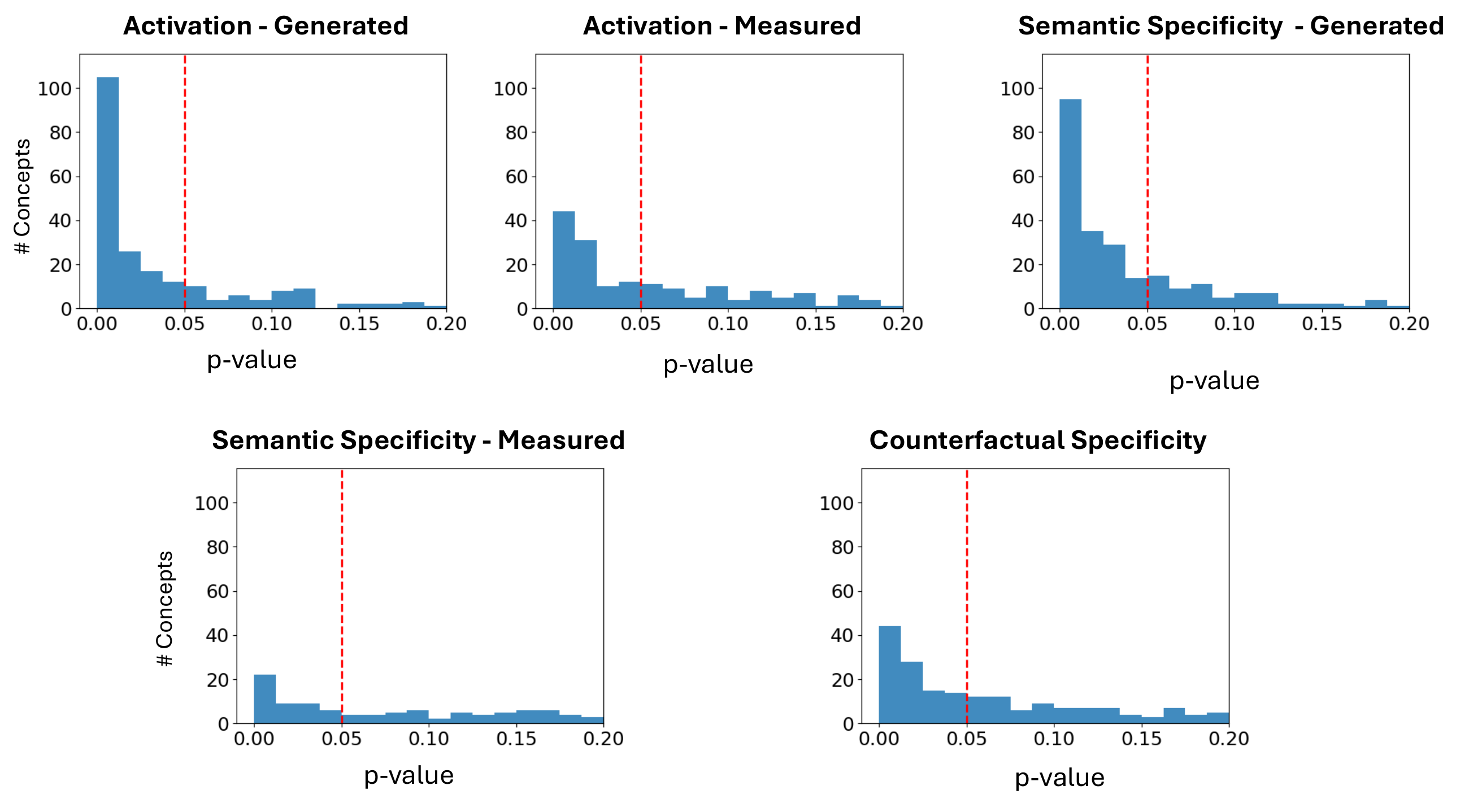}
    \caption{\textbf{BrainTRACE significance test.}
    For each discovered region, we compare the target concept both activation and specificity scores against scores obtained from other concepts on the same region. Histograms show one-sided empirical $p$-values for the five validation criteria. The red dashed line marks $p=0.05$. All p-values are computed on held-out validation data, after the voxel region has been selected on the training split.}
    \label{fig:baseline_pvalue_histograms}
\end{figure}

To further assess whether the discovered regions are selective for their corresponding target concepts, we performed a concept-specific baseline test. For each concept, we fixed the region discovered by the ranking method using $K=100$ voxels, and compared the region score for the target concept against the scores obtained by applying the same region to a set of concept-specific baseline concepts. The baseline set for each concept was defined by an LLM  to include only distinct concepts from the target. For each validation criterion, we computed a one-sided empirical $p$-value,
\[
p=\frac{1+\#\{s_{\mathrm{baseline}} \geq s_{\mathrm{target}}\}}{1+N_{\mathrm{baseline}}},
\]
testing whether the target concept score was higher than the baseline distribution. Figure~\ref{fig:baseline_pvalue_histograms} shows the resulting $p$-value distributions for the five validation criteria: Activation--Generated, Activation--Measured, Semantic Specificity--Generated, Semantic Specificity--Measured, and Counterfactual Specificity. Out of 260 tested concept--subject pairs, 160 passed Activation--Generated, 97 passed Activation--Measured, 173 passed Semantic Specificity--Generated, 47 passed Semantic Specificity--Measured, and 101 passed Counterfactual Specificity at $p \leq 0.05$. When considering only the generated-data activation and specificity criteria, 77 concept--subject pairs passed both tests. Requiring all five criteria to pass is conservative and yields a lower number of high-confidence discoveries. This does not imply that the remaining concepts are false; rather, many lack sufficient measured-data coverage for stringent validation. Accordingly, BrainTRACE separates high-confidence discoveries from promising but under-supported candidates that require follow-up stimuli.

\clearpage
\section{Additional Visual Results}

\begin{figure}[h]
    \centering
    \includegraphics[width=1.0\textwidth]{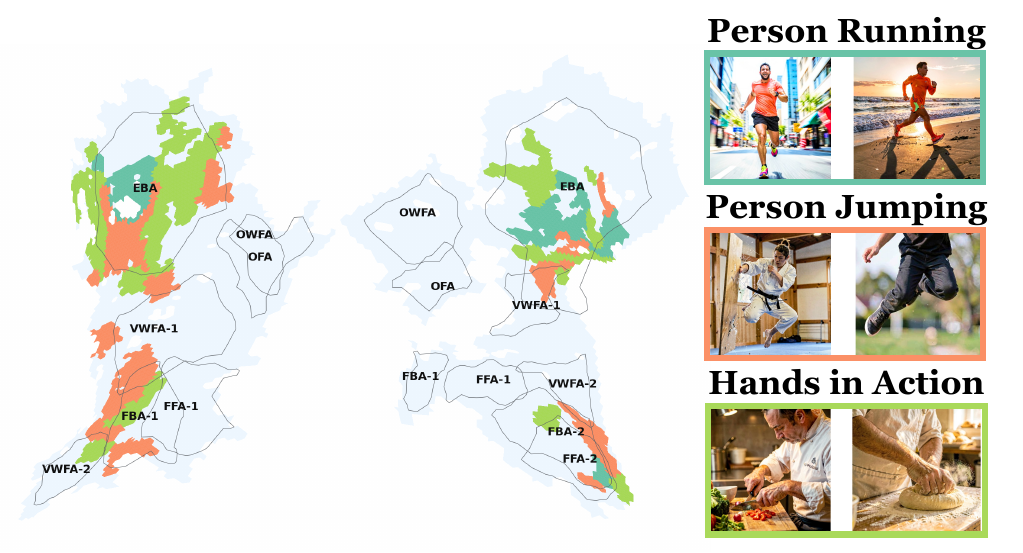}
    \caption{\textbf{Fine-grained organization of action- and body-related representations.} Binary maps show voxels with positive specificity evidence for Person Running, Person Jumping, and Hands in Action, with matching example images shown at right. BrainTRACE reveals partially distinct spatial patterns for these related body-action concepts across high-level visual cortex.}
    \label{sup_fig:Clusters_and_brains1}
\end{figure}

\begin{figure}[h]
    \centering
    \includegraphics[width=1.0\textwidth]{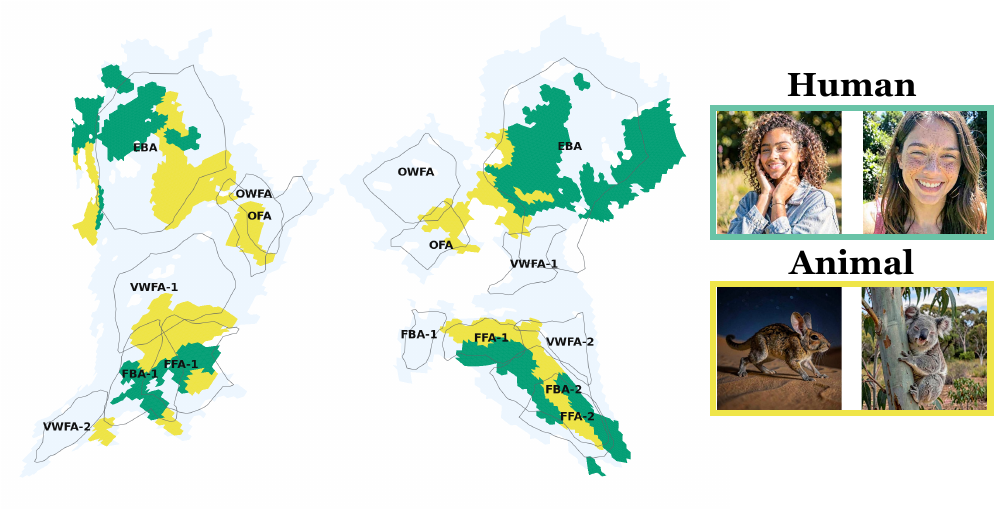}
    \caption{\textbf{Human and animal representations.} Binary maps show voxels with positive specificity evidence for Human and Animal, with matching example images shown at right. BrainTRACE reveals overlapping but distinct spatial patterns for these concepts across high-level visual cortex.}
    \label{sup_fig:Clusters_and_brains2}
\end{figure}

\clearpage
\section{Additional Technical Details}
\label{sec_sup:Additional_Details}

\subsection{Brain Data}
\label{sec_sup:fMRI data}

Functional magnetic resonance imaging (fMRI) measures changes in blood oxygenation that are related to neural activity over time. In visual fMRI experiments, subjects view images while whole-brain activity is recorded as a temporal sequence of 3D brain volumes. Following standard practice, this temporal signal is converted into image-level response estimates by fitting a response model, producing a beta value for each voxel and each presented image. Thus, after preprocessing, each image is associated with a single fMRI response vector, where each element corresponds to the estimated response of one voxel. We use the preprocessed fMRI responses provided by the Natural Scenes Dataset (NSD)~\cite{allen2022massive}, following the preprocessing and vectorization protocol used in prior work. The NSD contains 7T fMRI recordings from eight subjects viewing natural images from COCO, with each subject viewing approximately 10,000 images over repeated scanning sessions.  We further normalize each voxel independently across all measured images to have mean 0 and standard deviation 1, so that a response above 0 indicates activation above that voxel's average response across images.

\vspace{-0.15cm}
\paragraph{Large Predicted fMRI Pool.}
In addition to measured data, following the approach of \citet{Bao2025MindSimulator}, we construct a large pool of predicted image--fMRI pairs. Specifically, we take 120K unlabeled images from COCO and use the image-to-fMRI model to predict the corresponding fMRI response for each subject. These predicted pairs are used to support retrieval over a large image pool and to implement the MindSimulator baseline.

\subsection{Image \& Brain Models}
\label{sec_sup:image-brain_models}

\paragraph{Image-Based Models}
For both image generation and image editing, we use FLUX.2-Klein-4B from Black Forest Labs\footnote{\url{https://huggingface.co/black-forest-labs/FLUX.2-klein-4B}}.
For image generation, we use images of resolution $512 \times 512$, guidance scale $9.0$, and 35 inference steps. For image editing, we use the same resolution and guidance scale, but reduce the number of inference steps to 15.
For image--concept verification, we use Qwen3-VL-8B-Instruct\footnote{\url{https://huggingface.co/Qwen/Qwen3-VL-8B-Instruct}}.
Given an image and a target concept, the model is prompted to determine whether the concept is present in the image. We use temperature $0.0$ and constrain the model to answer only \texttt{yes} or \texttt{no}.

\vspace{-0.15cm}
\paragraph{Image-to-fMRI Prediction Model.}
Image-to-fMRI models are trained to predict a subject's fMRI response from a given image. We use the image-to-fMRI encoder from \citet{Beliy2024Wisdom}, which is trained jointly on image--fMRI pairs from all NSD subjects while still producing subject-specific predictions. This model allows us to take any image and predict the corresponding fMRI activation for each subject. We refer to these predicted responses as \emph{predicted fMRI}. Since fMRI prediction remains noisy, we filter out voxels whose encoding performance on a held-out test set is below a Pearson correlation of 0.2, reducing noise in the representation discovery process.

\subsection{Concept-Specific Image Retrieval}
\label{sec_sup:Image_Retrieval}

Following the MindSimulator approach, we use CLIP to retrieve images for a given concept. We first precompute CLIP image embeddings for all images in the NSD image--fMRI dataset, as well as for a large external pool of 120K images from COCO. Given a target concept, we encode it using the CLIP text encoder and compute the cosine similarity between the concept embedding and each image embedding, obtaining an alignment score for every image. Images are then ranked by this score, and the top-ranked images are retrieved. For the MindSimulator implementation used in our comparison, we use the same 120K external image pool, the same image-to-fMRI encoder, and retrieve the same number of images as generated by our method, in order to ensure a fair comparison. In practice, however, we found that many retrieved images do not clearly depict the desired concept, especially for more complex concepts where CLIP is less reliable and where COCO images often contain multiple objects and semantic elements. We therefore add a filtering step using a vision-language model (Qwen), which verifies whether the desired concept is indeed present in the retrieved image.

We also use retrieval from the measured image--fMRI dataset itself. For positive examples, retrieval is performed in the same way, followed by the same verification step to ensure that the target concept is clearly present. Retrieving semantic negatives is more challenging. For each target concept, the language model proposes a set of negative concepts, and for each such negative concept we seek measured images that satisfy two conditions simultaneously: the desired negative concept should be present, and the positive target concept should be absent. To enforce this, we use a two-stage CLIP-based retrieval procedure. First, we retrieve the top 100 images according to alignment with the negative concept. We then re-rank these candidates according to their distance from the positive concept, favoring images that are less aligned with the target concept. Finally, we apply a double verification step using the vision-language model: one check ensures that the negative concept is present, and the other ensures that the positive concept is absent. This retrieval pipeline allows us to obtain measured positive and semantic-negative examples that can be used for measured-data validation of the discovered representations.

\subsection{Extended Details on Concept-Selective Representation Search}
\label{sec_sup:Search}

Given a target concept, BrainTRACE uses the concept-targeted stimulus set to assign each voxel both an activation score and specificity scores. Let \(P=\{x_i^{+}\}_{i=1}^{N_{P}}\) denote the set of positive images, \(N=\{x_j^{-}\}_{j=1}^{N_{N}}\) the set of semantic negative images, and for each positive image \(x_i^{+}\), let \(E_i=\{x_{i,k}^{\mathrm{edit}}\}_{k=1}^{N_{E_i}}\) denote its corresponding counterfactual edited images. For a voxel \(v\), let \(a_v(x)\) denote its activation on the fMRI response associated with image \(x\), whether measured or predicted.

We first define the \emph{\textbf{Positive Score}} of voxel \(v\) as the average activation on the positive images,
\[
S_{\mathrm{pos}}(v)=\frac{1}{N_P}\sum_{i=1}^{N_P} a_v(x_i^{+}).
\]
This measures how strongly the voxel responds, on average, to the target concept. To test specificity against semantically related alternatives, we define the \emph{\textbf{Semantic-Negative Score}} by comparing the positive score to the hardest semantic negatives for that voxel. Specifically, let \(N_v^{\mathrm{hard}}\subseteq N\) be the set of the top 10 semantic negative images with the highest activations under voxel \(v\). We then define
\[
S_{\mathrm{neg}}(v)=
\frac{1}{N_P}\sum_{i=1}^{N_P} a_v(x_i^{+})
-
\frac{1}{|N_v^{\mathrm{hard}}|}\sum_{x\in N_v^{\mathrm{hard}}} a_v(x).
\]
A high value indicates that the voxel responds more strongly to the target concept than to the most confusable semantic alternatives. To test specificity under controlled edits, we define the \emph{\textbf{Counterfactual Score}} by comparing each positive image to its hardest edited version. For each positive image \(x_i^{+}\), let
\[
e_i^*(v)=\max_{x\in E_i} a_v(x),
\]
that is, the activation of voxel \(v\) on the single hardest edited negative associated with \(x_i^{+}\). The counterfactual specificity score is then
\[
S_{\mathrm{edit}}(v)=
\frac{1}{N_P}\sum_{i=1}^{N_P}\Big(a_v(x_i^{+})-e_i^*(v)\Big).
\]
A high value indicates that the voxel's response decreases when the target concept is specifically removed or replaced while the rest of the image is kept as similar as possible.

Using these scores, BrainTRACE constructs a candidate representation for the concept as a set of voxels whose average specificity score is positive. Concretely, for each voxel \(v\), we define its specificity score as the average of the semantic-negative and counterfactual specificity scores,
\[
S_{\mathrm{spec}}(v)=\frac{1}{2}\Big(S_{\mathrm{neg}}(v)+S_{\mathrm{edit}}(v)\Big),
\]
and select the voxel set
\[
R=\{v:\; S_{\mathrm{spec}}(v)>0\}.
\]
This set serves as the candidate region representing the concept. To evaluate the candidate region as a whole, BrainTRACE averages activations across voxels in \(R\) and computes the same scores at the region level. Specifically, for any image \(x\), the region activation is
\[
a_R(x)=\frac{1}{|R|}\sum_{v\in R} a_v(x),
\]
and the positive, semantic-negative specificity, and counterfactual specificity scores are computed by replacing \(a_v(\cdot)\) with \(a_R(\cdot)\) in the definitions above. The specificity score of the discovered region on the training set is used to determine whether the region is a strong candidate representation for the concept. Final quantitative evaluation is then performed on a separate evaluation split and on measured fMRI data, as described in Sec.~\ref{sec:method:Steps:Final}.

\subsection{Computational Resources}
\label{sec_sup:resources}

Our pipeline uses large models during the data creation process. We ran all experiments on H200 GPUs, although smaller GPUs can also be used when accessing the open-source models through APIs. All models used in our work are open source. Given a target concept to query, the full process of positive and negative image generation, yielding approximately 1000 images per concept, score computation, and region proposal takes about 2 hours on a single H200 GPU.


\end{document}